\documentclass{article}

\PassOptionsToPackage{numbers, compress}{natbib}

\usepackage[final]{neurips_2022}

\usepackage{amsmath}
\usepackage{amssymb}
\usepackage{mathtools}
\usepackage{amsthm}

\usepackage{hyperref}       % hyperlinks
\usepackage{url}            % simple URL typesetting
\usepackage{booktabs}       % professional-quality tables
\usepackage{amsfonts}       % blackboard math symbols
\usepackage{nicefrac}       % compact symbols for 1/2, etc.
\usepackage{microtype}      % microtypography
\usepackage{xcolor}         % colors
\usepackage{graphicx}
\usepackage{times}
\usepackage{latexsym}
\usepackage{multirow}
\usepackage{balance}
\usepackage{bbm}
\usepackage{paralist}
\usepackage{makecell}
\usepackage{subcaption}
\usepackage{caption}
\usepackage{xspace}
\usepackage{multicol}
\usepackage{transparent}
\usepackage{array}
\usepackage{bm}
\usepackage{tabularx}
\usepackage{setspace}
\usepackage{wrapfig,lipsum,booktabs}
\usepackage{pifont}
\usepackage{algorithm2e}

\newcommand{\fullres}[2]{$#1_{#2}$}
\newcommand{\eg}{\textit{e.g.}}
\newcommand{\ie}{\emph{i.e.}}
\newcommand{\method}{SuperGen\xspace}
\newcommand{\bs}[1]{\boldsymbol{#1}}

\title{Generating Training Data with Language Models: Towards Zero-Shot Language Understanding}

\author{%
  Yu Meng, Jiaxin Huang, Yu Zhang, Jiawei Han \\
  Department of Computer Science, University of Illinois at Urbana-Champaign \\
  \texttt{\{yumeng5,jiaxinh3,yuz9,hanj\}@illinois.edu} \\
}

\begin{document}

\maketitle

\begin{spacing}{0.96}

\begin{abstract}

Pretrained language models (PLMs) have demonstrated remarkable performance in various natural language processing tasks: Unidirectional PLMs (\eg, GPT) are well known for their superior text generation capabilities; bidirectional PLMs (\eg, BERT) have been the prominent choice for natural language understanding (NLU) tasks. 
While both types of models have achieved promising few-shot learning performance, their potential for zero-shot learning has been underexplored. 
In this paper, we present a simple approach that uses both types of PLMs for fully zero-shot learning of NLU tasks without requiring any task-specific data: A unidirectional PLM generates class-conditioned texts guided by prompts, which are used as the training data for fine-tuning a bidirectional PLM. 
% Our approach is compatible with any fine-tuning method.
With quality training data selected based on the generation probability and regularization techniques (label smoothing and temporal ensembling) applied to the fine-tuning stage for better generalization and stability, our approach demonstrates strong performance across seven classification tasks of the GLUE benchmark (\eg, $72.3$/$73.8$ on MNLI-m/mm and $92.8$ on SST-2), significantly outperforming zero-shot prompting methods and achieving even comparable results to strong few-shot approaches using $32$ training samples per class\footnote{Code can be found at \url{https://github.com/yumeng5/SuperGen}.}.

\end{abstract}

\section{Introduction}
Pretrained language models (PLMs)~\cite{Brown2020LanguageMA,clark2020electra,devlin2019bert,he2020deberta,liu2019roberta,meng2021coco,Meng2022PretrainingTE} have achieved human-level performance on natural language understanding (NLU) tasks~\cite{wang2018glue,Wang2019SuperGLUEAS} when fine-tuned on a large amount of task-specific training data.
However, such a supervised fine-tuning paradigm is drastically different from how humans perform these tasks: We barely need to see many task-specific training samples to perform well. 
Recently, many studies have revealed the intriguing few-shot learning potential  of PLMs: By converting task descriptions to natural language prompts and injecting them into PLMs, prompt-based approaches~\cite{Brown2020LanguageMA,gao2021making,Scao2021HowMD,Schick2021ExploitingCF,Schick2021ItsNJ} leverage task-specific information for better training data efficiency and have achieved remarkable few-shot results.

When prompt-based methods are applied to the zero-shot setting, however, the PLMs' predictions are much less accurate. 
For example, GPT-3's zero-shot performance is much degraded relative to its few-shot performance~\cite{Brown2020LanguageMA}, especially on challenging tasks like natural language inference (NLI). 
Without any task-specific samples, it is indeed challenging for PLMs to effectively interpret the prompts that come in different formats and are unseen in the pretraining data.
To familiarize PLMs with various prompts for zero-shot generalization to unseen tasks, a recent study proposes instruction tuning~\cite{Wei2022FinetunedLM}, which fine-tunes PLMs on a large collection of different tasks described by instructions. 
Despite its strong performance, its success is grounded in the large number of cross-task annotated datasets (\eg, train on many non-NLI tasks and transfer to NLI tasks) and the gigantic model size (\eg, hundreds of billions of parameters), posing great challenges for training and using them.
% in most applications.

In this work, we study zero-shot learning of PLMs on NLU tasks without any task-specific or cross-task data.
% \yuz{since COCO-LM is used in your major experiments, better say ``whose number of parameters is on par with RoBERTa$_{\text{Large}}$''?}
Motivated by the strong text generation power of recent PLMs~\cite{Brown2020LanguageMA,Keskar2019CTRLAC,Lewis2020BARTDS,raffel2019t5}, we propose \method, a \textbf{Super}vision \textbf{Gen}eration approach, wherein training data are created via a unidirectional PLM (\ie, the generator) which generates class-conditioned texts guided by label-descriptive prompts. 
A bidirectional PLM (\ie, the classifier) is then fine-tuned on the generated texts to perform the corresponding task.
Both PLMs can be of moderate size to fit in typical research hardware (\eg, a GPT-2-sized~\cite{radford2019language} generator and a RoBERTa$_{\text{Large}}$-sized~\cite{liu2019roberta} classifier). 
% \yuz{better say both uniPLM and biPLM are moderate sized? CTRL's number of parameters is also on par with GPT-2.}
With supervision automatically created by the generator, \method eliminates the need for task-specific annotations and provides the classifier PLM with a larger amount of training data than in few-shot scenarios.
% \method is compatible with any PLM as the classifier and any fine-tuning method.
% We note that the generator creates synthetic samples in a zero-shot manner, and the classifier is fine-tuned on the synthetic data (the classifier is thus not zero-shot, but there are no task-specific data required in such a process).
We call such a setting zero-shot because the entire process does not need any human annotated data, either from the target task or other tasks. 
The major difference from previous methods is that we synthesize training data for the target task, whereas existing zeros-shot methods do not use any form of training data from the test domain (but may train on other domains) and directly perform inference on the target task.

Across seven classification tasks of the GLUE benchmark~\cite{wang2018glue}, \method significantly outperforms the prompt-based zero-shot method and even achieves an overall better result in both average performance and stability than strong few-shot approaches that use $32$ annotated samples per class.
We identify several key factors to the strong performance of \method through ablation studies: (1) selecting quality training data based on their generated probability, and (2) using label smoothing and temporal ensembling to regularize fine-tuning on generated data.

\section{Related Work}

\subsection{Few-Shot and Zero-Shot Learning with PLMs}
Instead of using a large amount of annotated training data for fine-tuning PLMs on downstream tasks, few-shot learning studies how to better leverage only a small amount of task-specific training data, a more realistic scenario in many applications.
The most strict few-shot learning setting does not assume access to any unlabeled data or large validation sets for hyperparameter tuning~\cite{Perez2021TrueFL}, where prompt-based methods~\cite{Brown2020LanguageMA,gao2021making,Liu2021GPTUT,logan2021cutting,Scao2021HowMD,Schick2021ExploitingCF,Schick2021FewShotTG,Schick2021ItsNJ,Tam2021ImprovingAS,zhang2022differentiable} are prominently deployed to inject task descriptions into PLMs and make effective use of their language modeling capability for improved training data efficiency in low-data regimes.
More broadly, semi-supervised learning additionally leverages unlabeled task-specific data, where data augmentation~\cite{Chen2020MixTextLI,Xie2020UnsupervisedDA},  regularization~\cite{Miyato2017AdversarialTM} and bootstrapping~\cite{Schick2021ExploitingCF} methods are commonly used.

Zero-shot learning, on the other hand, is a much more challenging setting with absolutely no access to any task-specific data.
When prompt-based methods are directly used to obtain predictions from PLMs without any training, their zero-shot performance can be much worse~\cite{Brown2020LanguageMA,gao2021making}---difficult NLU tasks can be barely formulated as prompts that resemble the format of pretraining data, posing great challenges for PLMs to accurately interpret and leverage the prompts without given any training samples.
The current mainstream of zero-shot learning is based on transfer learning: By converting a set of tasks with abundant annotations into instruction templates~\cite{mishra2021cross,Sanh2021MultitaskPT,Wei2022FinetunedLM,Xu2022ZeroPromptSP}, entailment pairs~\cite{Yin2019BenchmarkingZT,Yin2020UniversalNL} or question-answer formats~\cite{Puri2019ZeroshotTC,Zhong2021AdaptingLM} and fine-tuning PLMs on them, the PLMs acquire the cross-task transfer ability~\cite{Ye2021CrossFitAF} to execute unseen tasks when they are formulated in a similar format.
Our work proposes a different approach from these studies: We use a unidirectional PLM to generate training data for fine-tuning another PLM on the target task. This not only removes the need for a large amount of cross-task annotations, but also eliminates the task difference in training and inference. Moreover, different from previous studies~\cite{anaby2020not,yang2020generative} that rely on labeled data to fine-tune the generative PLM, we directly use prompts to guide data generation without fine-tuning.

\subsection{Controlled Text Generation with PLMs}
Controlled text generation~\cite{Hu2017TowardCG} aims to steer the generated texts of language models towards desired contents, styles or domains.
Through fine-tuning PLMs on attribute-specific data, high-level control (\eg, generating certain topics or sentiments~\cite{Ziegler2019FineTuningLM}), fine-grained control (\eg, generating specific words or phrases~\cite{Chan2021CoConAS}) or both~\cite{Khalifa2021ADA} can be achieved.
Adapting PLMs to generate texts of specific attributes can also be realized at inference time without any further training of the PLMs~\cite{Dathathri2020PlugAP,Krause2021GeDiGD,Kumar2021ControlledTG,Liu2021DExpertsDC,Pascual2021APM,Yang2021FUDGECT}.
Different text attributes can also be represented during pretraining time as control codes~\cite{Keskar2019CTRLAC} which later can serve as explicit guidance for generating domain/attribute-specific texts. 

The idea of generating category-conditioned texts as training data has been explored for topic classification with bag-of-words or LSTM-based language models~\cite{Meng2018WeaklySupervisedNT,Meng2019WeaklySupervisedHT}, which may not have enough capacity to generate quality training data for challenging NLU tasks.
With more powerful PLMs, the idea of using prompts as guidance has emerged recently: Since natural language generation is largely based on contexts, using certain prompts to start a sequence can effectively steer the subsequent texts to be generated. 
The prompts can be either in natural language~\cite{Schick2021FewShotTG} or as learnable parameters~\cite{Li2021PrefixTuningOC}.
In this work, we also guide text generation via prompts, but for the novel purpose of creating training data for NLU tasks.
There have been studies with similar goals, such as generating similar/dissimilar sentences for training sentence embeddings~\cite{Schick2021GeneratingDW} and using labeled samples as demonstrations to prompt large PLMs~\cite{Yoo2021GPT3MixLL} for creating novel training data.
In this work, we explore generating training data \emph{without using any labeled samples} for a wide range of different NLU tasks.
The similar setting is also explored in a concurrent study~\cite{Ye2022ZeroGenEZ}.
Compared to annotated task-specific data, the generated texts may contain noise and have domain difference from the downstream task.
We introduce several important strategies for effective fine-tuning on generated data.

\begin{figure*}[!t]
% \subfigcapmargin=10pt
\centering
\includegraphics[width=1.0\textwidth]{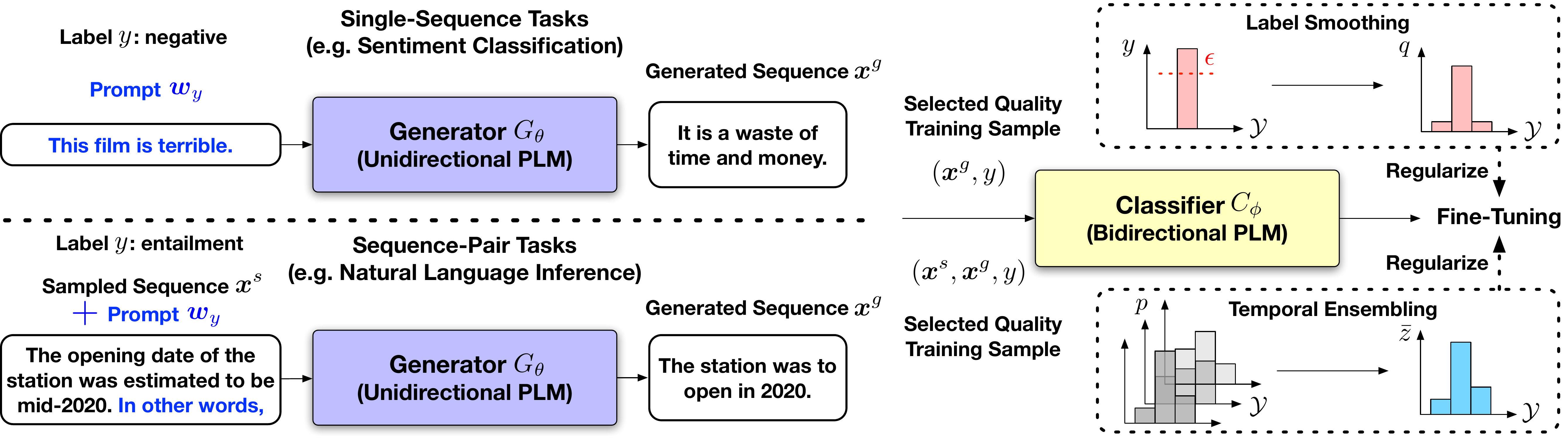}
\caption{Overview of \method for zero-shot learning of NLU tasks. A unidirectional PLM generates training data guided by label-descriptive prompts. Quality training samples are selected based on average log generation probability. A bidirectional PLM is fine-tuned on the selected training set with label smoothing and temporal ensembling as regularization to perform the classification task.}
\label{fig:overview}
\vspace{-1em}
\end{figure*}

\section{Method}
\subsection{Preliminaries}
\paragraph{Problem Formulation.} We consider solving a classification problem\footnote{We do not consider regression tasks in this work due to the difficulty of generating texts conditioned on a continuous label space. However, there exist approaches~\cite{Gao2021SimCSESC,Reimers2019SentenceBERTSE} that solve regression tasks by training on classification tasks. We leave the integration of \method with these methods as future work for solving regression tasks.} where we are only given the label space $\mathcal{Y}$ and a mapping $\mathcal{M}: \mathcal{Y} \to \mathcal{W}$ that converts each label $y \in \mathcal{Y}$ into a label-descriptive prompt (\ie, a short phrase) $\bs{w}_y \in \mathcal{W}$.
% We aim to train a model that classifies unseen texts into $\mathcal{Y}$.
We assume access to a unidirectional PLM $G_{\theta}$ as the generator and a bidirectional PLM $C_{\phi}$ which will be fine-tuned as the classifier\footnote{We assume the classifier to be bidirectional PLMs since they generally work better than unidirectional PLMs in NLU tasks; we can in principle use any PLM as the classifier.}.
We also assume the pretraining corpus $\mathcal{D}$ (\eg, Wikipedia) is available.
Fig.~\ref{fig:overview} shows an overview of our proposed \method method.
\paragraph{Text Generation with Unidirectional PLMs.}
A unidirectional PLM $G_{\theta}$ is pretrained to maximize the generation probability of each token in a sequence $\bs{x} = [x_1, x_2, \dots, x_n]$ conditioned on previous tokens:
$$
\max_{\theta} \prod_{i=1}^n p_{\theta}(x_i|\bs{x}_{<i}), \ \ \ \ {\rm where} \ \ \ \ p_{\theta}(x_i|\bs{x}_{<i}) = \frac{\exp(\bs{e}_i^\top \bs{h}_i)}{\sum_{j=1}^{|V|}\exp(\bs{e}_j^\top \bs{h}_i)}.
$$
Here, $p_{\theta}(\cdot)$ is usually parameterized using token embeddings $\bs{e}$ and contextualized embeddings $\bs{h}$ given by a Transformer~\cite{vaswani2017attention} encoder.
% where $p_{\theta}(\cdot)$ is usually parameterized using token embeddings $\bs{e}$ and contextualized embeddings $\bs{h}$ given by a Transformer~\cite{vaswani2017attention} encoder:
% $$
% p_{\theta}(x_i|\bs{x}_{<i}) = \frac{\exp(\bs{e}_i^\top \bs{h}_i)}{\sum_{j=1}^{|V|}\exp(\bs{e}_j^\top \bs{h}_i)}.
% $$

After pretraining, $G_{\theta}$ can be directly used to generate new texts by recursively sampling tokens from its output probability distribution. Typically, a temperature hyperparameter $\tau > 0$ is introduced during sampling~\cite{Hinton2015DistillingTK} to adjust the sharpness of the probability distribution:
% \yuz{any reference of using $\tau$ in generation?}
\begin{equation}
\label{eq:temp_prob}
p_{\theta}(x_i|\bs{x}_{<i}) = \frac{\exp(\bs{e}_i^\top \bs{h}_i/\tau)}{\sum_{j=1}^{|V|}\exp(\bs{e}_j^\top \bs{h}_i/\tau)},
\end{equation}
where $\tau \to 0$ approximates greedily picking the most probable next token; $\tau \to \infty$ induces a uniform distribution.
Additionally, sampled tokens can be confined to the top-$k$ most probable ones to avoid low-quality tokens.
In this work, we find such top-$k$ sampling with temperature is sufficient to produce coherent and meaningful texts as training data for NLU tasks. 
Exploring more sophisticated sampling strategies~\cite{Holtzman2020TheCC} is left for future work.

\begin{wraptable}{r}{7.5cm}
\caption{
Prompts used to generate class-conditioned texts for different GLUE tasks.
SST-2 is a single-sequence classification task and the rest are sequence-pair classification tasks. 
Generation for CoLA does not use prompts but by varying sampling temperatures.
$\bs{x}^s$ denotes a sequence randomly sampled from the pretraining corpus; $\bs{x}^g$ denotes the sequence to be generated by $G_{\theta}$; $\dots$ denotes skipping at least one sequence. 
% A few labels have two prompts and we show one of them for clarity.
See Appendix~\ref{app:prompts} for more details.
% \yuz{Better explain why the prompt of QNLI is different from that of MNLI and RTE. Better put SST-2 at the top, marking it as ``generating single sequences''; marking the other datasets as ``generating sequence pairs''. Better reorder the datasets according to the tasks?}
}
\centering
\small 
\resizebox{0.5\columnwidth}{!}{
\begin{tabular}{lll}
\toprule
\textbf{Task} & \textbf{Label} & \textbf{Prompt} \\
\midrule
\multirow{2}{*}{\textbf{SST-2}} & positive & Rating: 5.0 $\bs{x}^g$ \\
& negative & Rating: 1.0 $\bs{x}^g$ \\
\midrule
\multirow{4}{*}{\textbf{MNLI}} & entailment & $\bs{x}^s$. In other words, $\bs{x}^g$ \\
& neutral & $\bs{x}^s$. Furthermore, $\bs{x}^g$ \\
& \multirow{2}{*}{contradiction} & There is a rumor that $\bs{x}^s$. \\
& & However, the truth is: $\bs{x}^g$ \\
\midrule
\multirow{2}{*}{\textbf{QNLI}} & entailment & $\bs{x}^s$? $\bs{x}^g$ \\
& not entailment & $\bs{x}^s$? \dots $\bs{x}^g$ \\
\midrule
\multirow{2}{*}{\textbf{RTE}} & entailment & $\bs{x}^s$. In other words, $\bs{x}^g$ \\
& not entailment & $\bs{x}^s$. Furthermore, $\bs{x}^g$ \\
\midrule
\multirow{2}{*}{\textbf{MRPC}} & equivalent & $\bs{x}^s$. In other words, $\bs{x}^g$ \\
& not equivalent & $\bs{x}^s$. Furthermore, $\bs{x}^g$ \\
\midrule
\multirow{2}{*}{\textbf{QQP}} & equivalent & $\bs{x}^s$? In other words, $\bs{x}^g$ \\
& not equivalent & $\bs{x}^s$? Furthermore, $\bs{x}^g$ \\
\bottomrule
\end{tabular}
}
\vspace{-1em}
\label{tab:prompts}
\end{wraptable}

\subsection{Training Data Generation}
When given a label-descriptive prompt such as ``Write a negative review:'', humans are able to produce texts pertaining to the corresponding class.
We aim to leverage the strong text generation power of a unidirectional PLM $G_{\theta}$ for the same purpose of creating class-conditioned training data.
We note that $G_{\theta}$ is directly used for generation without any parameter updates. 
The prompts used for different NLU tasks in GLUE are summarized in Table~\ref{tab:prompts}.

\paragraph{Generating Single Sequences.}
For single-sequence NLU tasks such as sentiment classification (\eg, SST-2), we simply use a prompt $\bs{w}_y$ corresponding to label $y$ as the beginning of the sequence and let $G_{\theta}$ generate the remaining sequence:
$$
\bs{x}^g \gets G_{\theta}(\bs{w}_y),
$$
where $G_{\theta}(\bs{w}_y)$ denotes using $\bs{w}_y$ as the input to $G_{\theta}$ and recursively sampling tokens from the distribution in Eq.~\eqref{eq:temp_prob} until a full sequence is generated; $\bs{x}^g$ denotes the \emph{generated} sequence (\ie, excluding the prompt), which will be paired with $y$ to form one training sample $(\bs{x}^g, y)$.

For syntactic tasks like linguistic acceptability classification (\eg, CoLA) which requires generating both linguistically acceptable and unacceptable sequences, we start the sequence with random stop words and use varying sampling temperatures for generating different sequences.
A smaller temperature (\eg, $\tau=0.1$ in Equation~\eqref{eq:temp_prob}) sharpens the sampling probability distribution towards the most probable tokens, thus the resulting sequence is more likely to be linguistically acceptable.
Using a larger temperature (\eg, $\tau=10$ in Equation~\eqref{eq:temp_prob}) flattens the sampling probability distribution to be more uniform, and the generated tokens will be nearly random, which can create linguistically incorrect sequences.

\paragraph{Generating Sequence Pairs.}
Sequence-pair classification tasks require generating two sequences of specific relationships (\eg, entailment, contradiction). 
We sample\footnote{In principle, we can also generate the first sequence using $G_{\theta}$, but we find sampling from $\mathcal{D}$ improves the diversity of texts.} the first sequence $\bs{x}^s$ from the pretraining corpus $\mathcal{D}$, concatenate the prompt $\bs{w}_y$ with $\bs{x}^s$, and generate the second sequence $\bs{x}^g$:
$$
\bs{x}^g \gets G_{\theta}\left([\bs{x}^s;\bs{w}_y]\right),\, \bs{x}^s \sim \mathcal{D}.
$$
The sequence pair training sample will then be formed as $(\bs{x}^s, \bs{x}^g, y)$.

\paragraph{Rewarding and Penalizing Repetitions for Sequence Pair Generation.}
A common issue in text generation is degenerate repetition~\cite{Holtzman2020TheCC,Keskar2019CTRLAC,radford2019language,Welleck2020NeuralTG} where generated texts get stuck in repetition loops.
To address this issue, one approach is to discourage repetition by reducing the logits of tokens that are already in the sequence before performing sampling~\cite{Keskar2019CTRLAC}. 
In sequence pair generation, however, it is sometimes desirable to encourage the second sequence to repeat some words in the first sentence (\eg, for generating an entailment or a paraphrase).
Therefore, we propose a simple modification of Eq.~\eqref{eq:temp_prob} that rewards/penalizes repetition based on whether the token has appeared in $\bs{x}^s$/$\bs{x}^g$:
\begin{equation}
\label{eq:temp_prob_mod}
p_{\theta}(x_i|\bs{x}_{<i}) = \frac{\exp(\bs{e}_i^\top \bs{h}_i/\omega)}{\sum_{j=1}^{|V|}\exp(\bs{e}_j^\top \bs{h}_i/\omega)}, \ \ \ \ {\rm where} \ \ \ \ \omega = \begin{cases}
\tau \alpha & x_i \in \bs{x}^s \land x_i \notin \bs{x}^g \\
\tau \beta & x_i \in \bs{x}^g \\
\tau & \text{else}
\end{cases},
\end{equation}
% where 
% $$
% \omega = \begin{cases}
% \tau \alpha & x_i \in \bs{x}^s \land x_i \notin \bs{x}^g \\
% \tau \beta & x_i \in \bs{x}^g \\
% \tau & \text{else}
% \end{cases},
% $$
and $\alpha > 0, \beta > 0$ are hyperparameters. 
By setting $\alpha < 1$ and $\beta > 1$, we can promote tokens in $\bs{x}^s$ that have not appeared in $\bs{x}^g$ to have a higher chance of being generated, and discourage the generation of repetitive tokens in $\bs{x}^g$ to mitigate degenerate repetition.
The parameters used for different tasks are listed in Appendix~\ref{app:hyperpara} Table~\ref{tab:gen_hyperpara}.

\subsection{Effective Fine-Tuning on Generated Texts}
\label{sec:fine-tune}
With the generated training data, one can fine-tune a bidirectional PLM $C_{\phi}$ as the classifier to perform the NLU task.
However, training $C_{\phi}$ via standard supervised training on all generated texts is likely to yield suboptimal performance on downstream tasks because (1) the generated texts may contain noise as $G_{\theta}$ may not always produce texts pertaining to the desired class, especially for challenging sequence pair tasks with subtle semantic relationships; and (2) the generated texts can be considered as originated from the domain of $G_{\theta}$'s pretraining data, with a potentially different distribution from the downstream task; straightforward application of supervised training will result in overfitting to the pretraining domain and diminishing generalization ability, a common challenge in transfer learning~\cite{torrey2010transfer,zhuang2020comprehensive}.
To address these challenges, we next introduce several simple and important strategies for more effective and stable fine-tuning on generated texts.

\paragraph{Selecting Quality Training Data.}
We aim to select generated texts $\bs{x}^g$ that are most likely to pertain to the desired label $y$ (\ie, with the highest $p(\bs{x}^g|y)$). 
% By assuming a uniform prior $p(y)$ over all labels $y \in \mathcal{Y}$, we have 
% \yuz{why can you assume the uniform prior? any statistics from GLUE?}
% $$
% p(y|\bs{x}^g) \propto p(\bs{x}^g|y)p(y) \stackrel{\text{uniform prior}}{\propto} p(\bs{x}^g|y).
% $$
The true probability $p(\bs{x}^g|y)$ is unknown and we estimate it via the generation probability given by $G_{\theta}$ conditioned on the prompt $\bs{w}_y$:
$$
p(\bs{x}^g|y) \approx p_{\theta}(\bs{x}^g|\bs{w}_y) = \prod_{i=1}^n p_{\theta}\left(x_i \big| [\bs{w}_y;\bs{x}^g_{<i}]\right).
$$
Since the above measure is biased towards shorter sequences, we instead use the geometric mean of the above conditional generation probability (or equivalently, the average log probability) of all tokens in $\bs{x}^g$ as the ranking score, following~\cite{Yuan2021BARTScoreEG}:
\begin{equation}
\label{eq:sorting_score}
r = \frac{1}{n} \sum_{i=1}^n \log p_{\theta}\left(x_i \big| [\bs{w}_y;\bs{x}^g_{<i}]\right).
\end{equation}
To construct a training set consisting of $N$ samples per class, we will generate more samples (\eg, $10N$), and select training data based on the score $r$ in Eq.~\eqref{eq:sorting_score}: 
For all tasks except CoLA, the top-$N$ ones of each class are selected; for CoLA, the top-$N$ ones are used as linguistically acceptable training samples, and the bottom-$N$ ones as linguistically unacceptable sequences.

\paragraph{Regularization for Better Generalization and Stability.}
Even with the above training data selection procedure, the resulting training set may still contain noise and there exists domain difference from the downstream tasks.
We apply two regularization techniques, \emph{label smoothing}~\cite{Szegedy2016RethinkingTI} and \emph{temporal ensembling}~\cite{Laine2017TemporalEF} for better fine-tuning stability and generalization.
% to downstream tasks.

Given a training sample $(\bs{x}^g, y)$, \emph{label smoothing} trains the classifier $C_{\phi}$ to minimize the standard cross-entropy loss between the label and the classifier's prediction $p_{\phi}(\bs{x}^g)$, except that the label is a weighted average of the one-hot vector and a uniform distribution over all labels:
\begin{equation}
\label{eq:label_smooth}
\min_{\phi} -\sum_{j=1}^{|\mathcal{Y}|} q_j \log(p_{\phi}(\bs{x}^g)_j), 
\end{equation}
where $q_j = \mathbbm{1}(j = y)(1-\epsilon) + \epsilon/|\mathcal{Y}|$ and $\epsilon$ is the smoothing weight.
By forcing the classifier to be less confident on training data, label smoothing improves robustness to label noise~\cite{Lukasik2020DoesLS} and prevents overfitting to the training set~\cite{Mller2019WhenDL}, thus improving generalization to different domains.

The motivation for \emph{temporal ensembling} is that neural networks usually first pick up easy and general patterns in the data before learning more sophisticated and dataset-specific features~\cite{Zhang2017UnderstandingDL}, and thus the earlier states of the network offer better generalizability to different domains. We therefore record the predictions $\bs{p}_{\phi} = p_{\phi}(\bs{x}^g)$ of $C_{\phi}$ on each training sample $(\bs{x}^g, y)$ at different training steps, and use the accumulated moving-average predictions $\bar{\bs{z}}$ to regularize the latest model training. 
This also helps suppress the fluctuation in model predictions due to data noise, offering better noise-robustness~\cite{Nguyen2020SELFLT}.
We update ensembled predictions $\bar{\bs{z}}$ once every $B$ batches:
\begin{equation}
\label{eq:udpate_ens}
\hat{\bs{z}} \gets \gamma\hat{\bs{z}} + (1-\gamma)\bs{p}_{\phi}, \, \bar{\bs{z}} \gets \hat{\bs{z}}/(1-\gamma^t),
\end{equation}
where $\hat{\bs{z}}$ has a zero initialization; $\gamma$ is the momentum parameter; $t$ is the number of updates $\bar{\bs{z}}$ has received; the division $(1-\gamma^t)$ is for bias correction~\cite{Laine2017TemporalEF}. 
We also use the ensembled prediction $\bar{\bs{z}}$ as a reliable signal to filter out noisy training samples: Only those samples on which $\bar{\bs{z}}$ strongly agrees with the label $y$ (\ie, $\bar{z}_{y} > \delta$ where $\delta>0$ is a threshold parameter) will be used for training.

We regularize model training by extending Eq.~\eqref{eq:label_smooth} to add a KL divergence regularization term from the model prediction to the ensembled prediction weighed by $\lambda$:
\begin{equation}
\label{eq:train_obj}
\min_{\phi} -\sum_{j=1}^{|\mathcal{Y}|} q_j \log(p_{\phi}(\bs{x}^g)_j) - \lambda \sum_{j=1}^{|\mathcal{Y}|} \bar{z}_j \log \frac{p_{\phi}(\bs{x}^g)_j}{\bar{z}_j}.
\end{equation}
We follow \cite{Laine2017TemporalEF} to slowly ramp-up $\lambda$ during training.

\RestyleAlgo{ruled}
\SetKwInput{KwInput}{Input}
\SetKwInput{KwParameter}{Parameter}
\SetKwInput{KwOutput}{Output}
\newcommand\mycommfont[1]{\footnotesize\ttfamily\textcolor{blue}{#1}}
\SetCommentSty{mycommfont}

\begin{wrapfigure}[31]{r}{7.8cm}
\small
    \begin{algorithm}[H]
    \DontPrintSemicolon
    \SetNoFillComment
      \KwInput{$\mathcal{Y}$: Label space; $\mathcal{P}$: Label-descriptive prompts; $G_{\theta}$: Unidirectional PLM; $C_{\phi}$: Bidirectional PLM.}
      \KwParameter{$N$: Number of training samples per class to generate; $M (\gg N)$: Number of total training samples to generate; $T$: Number of training steps; $B$: Ensemble prediction update interval; $\delta$: Threshold parameter.}
      \KwOutput{$C^*_{\phi}$: Classifier that classifies input texts into $\mathcal{Y}$.}
      
      \For{$y \in \mathcal{Y}$}    
        { 
            $\mathcal{T}_y \gets \{\}$\; \tcp*[l]{Class $y$ train set init.}
            \For{$i \in [1, 2, \dots, M]$}
            {
                $\bs{x}^g \gets G_{\theta}(\bs{w}_y)$\; 
                
                % \tcp*[l]{Single sequences.}
                $\mathcal{T}_y \gets \mathcal{T}_y \bigcup \{(\bs{x}^g, y)\}$\;
            }
        	
        }
      $\mathcal{T} \gets \{\}$\; \tcp*[l]{Selected train set.}
      \For{$y \in \mathcal{Y}$}    
        { 
            Sort $\mathcal{T}_y$ in descending order by Eq.~\eqref{eq:sorting_score}\;
            
            $\mathcal{T} \gets \mathcal{T} \bigcup \mathcal{T}_y[:N]$\;
        }
    
      $\hat{\bs{z}} \gets \bs{0}$\; \tcp*[l]{Ensembled prediction init.}
      
      $\mathcal{T}^* \gets \mathcal{T}$\; \tcp*[l]{Filtered train set.}
      
        \For{$i \in [1, 2, \dots, T]$}    
        { 
            Fine-tune $C_{\phi}$ via Eq.~\eqref{eq:train_obj} on a minibatch of $\mathcal{T}^*$\;
            
            \If{$i \% B = 0$}
            {
                Update $\hat{\bs{z}}, \bar{\bs{z}}$ via Eq.~\eqref{eq:udpate_ens}\;
                
                $\mathcal{T}^* \gets \{(\bs{x}^g, y)|\bar{z}_{y} > \delta, (\bs{x}^g, y) \in \mathcal{T}\}$\;
            }
            
        }
        \Return $C^*_{\phi} = C_{\phi}$\;
    \caption{\method for Zero-Shot Learning.}
    \label{alg:main_alg}
    \end{algorithm}
\end{wrapfigure}
\subsection{Overall Algorithm}
We summarize \method for single-sequence NLU tasks in Algorithm~\ref{alg:main_alg}. Solving sequence-pair problems follows the same algorithm except the pretraining corpus $\mathcal{D}$ is needed for sampling the first sequence $\bs{x}^s$.

\section{Experimental Setup}

\paragraph{Downstream Tasks and Metrics.} 
We use all the tasks included in GLUE~\cite{wang2018glue} except STS-B which is a regression task.
Please refer to Appendix~\ref{app:glue} for more details about GLUE tasks.
We follow the evaluation protocol of \cite{gao2021making}: We use F1 score as the metric for QQP and MRPC, Matthews correlation for CoLA, and accuracy for the rest of the tasks.
The original development sets of these tasks are used for testing.
For all reported results, we include the average and standard deviation over $5$ different random seeds.
\paragraph{Models.} 
Unless specified otherwise, we use CTRL ($1.63$B parameters)~\cite{Keskar2019CTRLAC} as the generator $G_{\theta}$ and COCO-LM$_{\text{Large}}$ ($367$M parameters)~\cite{meng2021coco} as the classifier $C_{\phi}$. 
We also show the results using similar-sized PLMs (GPT-2~\cite{radford2019language}/RoBERTa~\cite{liu2019roberta}) as the generator/classifier in Section~\ref{sec:plms}. 
% \yuz{emphasize again their numbers of parameters are on par with GPT-2 and RoBERTa, respectievly?}
\paragraph{Fine-Tuning Settings and Hyperparameters.} 
We note that \method is compatible with any fine-tuning method; while using more sophisticated methods may grant further performance improvement, we use the basic prompt-based fine-tuning with manual templates approach for simplicity and clarity. 
For all tasks, we use the same templates and label words as in \cite{gao2021making}.
% For CoLA, we use standard fine-tuning since the input data might be out of the distribution of $C_{\phi}$~\cite{gao2021making}.
Under the zero-shot learning setting, it is not possible to tune hyperparameters due to the lack of validation sets. 
Therefore, we keep all fine-tuning hyperparameters (\eg, learning rate, batch size, training epochs, number of generated training samples, label smoothing and temporal ensembling hyperparameters) the same across all tasks.
See Appendix~\ref{app:hyperpara} Table~\ref{tab:finetune_hyperpara} for details.
\paragraph{Compared Methods and Ablations.} 
We include the results of zero-shot prompting, standard few-shot fine-tuning and the four few-shot prompt-based fine-tuning methods proposed in \cite{gao2021making}.
We also conduct ablation studies by removing the following three techniques from \method one at a time: (1) not using Eq.~\eqref{eq:sorting_score} for training data selection but randomly selecting the same amount of training data ($-$ data selection); (2) not using label smoothing ($-$ label smooth) but using one-hot labels; and (3) not using temporal ensembling (\ie, using Eq.~\eqref{eq:label_smooth} instead of Eq.~\eqref{eq:train_obj} as the training objective) ($-$ temporal ensemble).
Lastly, we include the fully supervised fine-tuning results trained on the entire training sets.

\begin{table}[!t]
\caption{
Results on seven GLUE classification tasks. We report average
and standard deviation (as subscripts) performance over $5$ different random seeds. $^\dagger$: Results from LM-BFF~\cite{gao2021making}.
}
\vspace{0.5em}
\centering
\small 
\resizebox{\textwidth}{!}{
\begin{tabular}{l*{8}{c}}
\toprule
\multirow{2}{*}{\textbf{Method}} & \textbf{MNLI-(m/mm)} & \textbf{QQP} & \textbf{QNLI} & \textbf{SST-2} & \textbf{CoLA} & \textbf{RTE} & \textbf{MRPC} & \textbf{AVG} \\ 
% \cmidrule(lr){2-3}
& (Acc.) & (F1) & (Acc.) & (Acc.) & (Matt.) & (Acc.) & (F1) &  \\
\midrule
\multicolumn{9}{l}{\textbf{Zero-Shot Setting:} No task-specific data (neither labeled nor unlabeled).}  \\ 
\midrule
Prompting$^\dagger$ & \fullres{50.8}{0.0}/\fullres{51.7}{0.0} & \fullres{49.7}{0.0} & \fullres{50.8}{0.0} & \fullres{83.6}{0.0} & \fullres{2.0}{0.0} & \fullres{51.3}{0.0} & \fullres{61.9}{0.0} & $50.1$ \\
\method & \fullres{\textbf{72.3}}{0.5}/\fullres{\textbf{73.8}}{0.5} & \fullres{\textbf{66.1}}{1.1} & \fullres{\textbf{73.3}}{1.9} & \fullres{\textbf{92.8}}{0.6} & \fullres{\textbf{32.7}}{5.5} & \fullres{\textbf{65.3}}{1.2} & \fullres{82.2}{0.5} & $\textbf{69.4}$ \\
\,\, $-$ data selection & \fullres{63.7}{1.5}/\fullres{64.2}{1.6} & \fullres{62.3}{2.2} & \fullres{63.9}{3.2} & \fullres{91.3}{2.0} & \fullres{30.5}{8.8} & \fullres{62.4}{1.5} & \fullres{81.6}{0.2} & $65.1$ \\
\,\, $-$ label smooth & \fullres{70.7}{0.8}/\fullres{72.1}{0.7} & \fullres{65.1}{0.9} & \fullres{71.4}{2.5} & \fullres{91.0}{0.9} & \fullres{9.5}{1.0} & \fullres{64.8}{1.1} & \fullres{\textbf{83.0}}{0.7} & $65.2$\\
\,\, $-$ temporal ensemble & \fullres{62.0}{4.6}/\fullres{63.6}{4.8} & \fullres{63.9}{0.3} & \fullres{72.4}{2.0} & \fullres{92.5}{0.9} &  \fullres{23.5}{7.0} & \fullres{63.5}{1.0} & \fullres{78.8}{2.2} & $65.3$\\
\midrule
\multicolumn{9}{l}{\textbf{Few-Shot Setting:} Use $32$ labeled samples/class (half for training and half for development).}  \\ 
\midrule
Fine-tuning$^\dagger$ & \fullres{45.8}{6.4}/\fullres{47.8}{6.8} & \fullres{60.7}{4.3} & \fullres{60.2}{6.5} & \fullres{81.4}{3.8} & \fullres{\textbf{33.9}}{14.3} & \fullres{54.4}{3.9} & \fullres{76.6}{2.5} & $59.1$ \\
Manual prompt$^\dagger$ & \fullres{68.3}{2.3}/\fullres{70.5}{1.9} & \fullres{65.5}{5.3} & \fullres{64.5}{4.2} & \fullres{92.7}{0.9} & \fullres{9.3}{7.3} & \fullres{69.1}{3.6} & \fullres{74.5}{5.3} & $63.6$ \\
\,\, $+$ demonstration$^\dagger$ & \fullres{\textbf{70.7}}{1.3}/\fullres{\textbf{72.0}}{1.2} & \fullres{\textbf{69.8}}{1.8} & \fullres{\textbf{69.2}}{1.9} & \fullres{92.6}{0.5} & \fullres{18.7}{8.8} & \fullres{68.7}{2.3} & \fullres{77.8}{2.0} & $66.9$ \\
Auto prompt$^\dagger$ & \fullres{68.3}{2.5}/\fullres{70.1}{2.6} & \fullres{67.0}{3.0} & \fullres{68.3}{7.4} & \fullres{92.3}{1.0} & \fullres{14.0}{14.1} & \fullres{\textbf{73.9}}{2.2} & \fullres{76.2}{2.3} & $65.8$ \\
\,\, $+$ demonstration$^\dagger$ & \fullres{70.0}{3.6}/\fullres{72.0}{3.1} & \fullres{67.7}{5.8} & \fullres{68.5}{5.4} & \fullres{\textbf{93.0}}{0.6} & \fullres{21.8}{15.9} & \fullres{71.1}{5.3} & \fullres{\textbf{78.1}}{3.4} & $\textbf{67.3}$ \\
\midrule
Fully supervised$^\dagger$ & \textit{89.8}/\textit{89.5} & \textit{81.7} & \textit{93.3} & \textit{95.0} & \textit{62.6} & \textit{80.9} & \textit{91.4} & $\textit{84.9}$ \\
\bottomrule
\end{tabular}
}
\label{tab:main_res}
\vspace{-1em}
\end{table}

\section{Evaluation}

\subsection{Main Results}
We present the results of \method, its ablations and compared methods in Table~\ref{tab:main_res}.
Overall, \method significantly outperforms zero-shot prompting and achieves an overall better result than all few-shot methods.
Notably, \method results in much smaller variance over different random seeds than few-shot approaches on most tasks---with access to more training data, fine-tuning of PLMs becomes much more stable.
The ablation results demonstrate that all three strategies (\ie, quality training data selection, label smoothing and temporal ensembling) play important roles in improving and stabilizing the final performance, especially on challenging tasks like MNLI.

\subsection{Using Different Prompts}
\label{sec:diff_prompts}
% \begin{table}[t]
% \caption{
% Results with different groups of prompts.
% Prompt group \#3 is not applicable to RTE/MRPC which do not have the contradiction label.
% }
% \vspace{-0.0em}
% \centering
% % \small 
% \resizebox{0.5\columnwidth}{!}{
% \begin{tabular}{l*{3}{c}}
% \toprule
% {\textbf{Prompt Group}} & \textbf{MNLI-(m/mm)} & \textbf{RTE} & \textbf{MRPC} \\ 
% \midrule
% \# 0 (Original) & \fullres{\textbf{72.3}}{0.5}/\fullres{\textbf{73.8}}{0.5} & \fullres{65.3}{1.2} & \fullres{\textbf{82.2}}{0.5} \\
% \# 1 & \fullres{70.7}{1.4}/\fullres{72.4}{1.2} & \fullres{64.4}{1.6} & \fullres{\textbf{82.2}}{0.2} \\
% \# 2 & \fullres{70.8}{0.6}/\fullres{72.1}{0.8} & \fullres{64.7}{1.8} & \fullres{81.8}{0.8} \\
% \# 3 & \fullres{70.9}{1.4}/\fullres{72.2}{1.4} & - & - \\
% Mixed & \fullres{72.2}{0.7}/\fullres{73.4}{0.6} & \fullres{\textbf{66.3}}{1.0} & \fullres{81.3}{2.0} \\
% \bottomrule
% \end{tabular}
% }
% \vspace{-1em}
% \label{tab:prompt_study}
% \end{table}

\begin{table}[t]
\centering

% \begin{minipage}{0.49\linewidth}
\centering
\caption{
Results with different groups of prompts.
% Prompt group \#3 is not applicable to RTE/MRPC which do not have the contradiction label.
CoLA does not use prompts for generation.
The number of prompt groups is equal to the number of the task labels.
}
\vspace{0.5em}
\small
% \resizebox{\columnwidth}{!}{
\begin{tabular}{l*{6}{c}}
\toprule
{\textbf{Prompt Group}} & \textbf{MNLI-(m/mm)} & \textbf{QQP} & \textbf{QNLI} & \textbf{SST-2} & \textbf{RTE} & \textbf{MRPC} \\ 
\midrule
$\#0$ (Original) & \fullres{\textbf{72.3}}{0.5}/\fullres{\textbf{73.8}}{0.5} & \fullres{66.1}{1.1} & \fullres{\textbf{73.3}}{1.9} & \fullres{\textbf{92.8}}{0.6} & \fullres{65.3}{1.2} & \fullres{\textbf{82.2}}{0.5} \\
$\#1$ & \fullres{70.7}{1.4}/\fullres{72.4}{1.2} & \fullres{65.5}{1.4} & \fullres{71.9}{1.7} & \fullres{92.2}{0.9} & \fullres{64.4}{1.6} & \fullres{81.9}{0.4} \\
$\#2$ & \fullres{70.8}{0.6}/\fullres{72.1}{0.8} & \fullres{65.6}{1.1} & \fullres{72.2}{2.2} & \fullres{92.4}{0.8} & \fullres{64.7}{1.8} & \fullres{81.8}{0.8} \\
$\#3$ & \fullres{70.9}{1.4}/\fullres{72.2}{1.4} & - & - & - & - & - \\
Mixed & \fullres{72.2}{0.7}/\fullres{73.4}{0.6} &  \fullres{\textbf{66.9}}{1.5} & \fullres{73.0}{1.7} & \fullres{\textbf{92.8}}{0.9} & \fullres{\textbf{66.3}}{1.0} & \fullres{81.3}{2.0} \\
\bottomrule
\end{tabular}
% }
\label{tab:prompt_study}
% \end{minipage}
% \hspace{1mm}
% \begin{minipage}{0.49\linewidth}
% \centering
% \caption{
% Results with different generator and classifier PLMs.
% }
% \vspace{0.5em}
% \resizebox{\columnwidth}{!}{
% \begin{tabular}{l*{2}{c}}
% \toprule
% {\textbf{PLMs}} & \textbf{MNLI-(m/mm)} & \textbf{SST-2} \\ 
% \midrule
% $G_{\theta}$: CTRL, $C_{\phi}$: COCO-LM & \fullres{\textbf{72.3}}{0.5}/\fullres{\textbf{73.8}}{0.5} &  \fullres{92.8}{0.6} \\
% $G_{\theta}$: CTRL, $C_{\phi}$: RoBERTa & \fullres{69.0}{0.8}/\fullres{70.6}{0.9} &  \fullres{\textbf{93.3}}{1.5} \\
% $G_{\theta}$: GPT-2, $C_{\phi}$: COCO-LM & \fullres{69.5}{1.2}/\fullres{71.3}{1.3} & \fullres{88.2}{1.8} \\
% $G_{\theta}$: GPT-2, $C_{\phi}$: RoBERTa & \fullres{68.3}{0.9}/\fullres{69.7}{0.7} & \fullres{88.6}{0.8} \\
% \bottomrule
% \end{tabular}
% }
% \label{tab:plm_ablation}
% \end{minipage}
% \vspace{-1.5em}
\end{table}

One important factor of \method is the choice of label-descriptive prompts as they directly influence the quality of generated training samples.
To study the impact of different prompt choices on the final model performance, we create different groups of prompts other than the original ones. 
We replace the prompt for one label used in Table~\ref{tab:prompts} with a synonymous one and keep other prompts unchanged when forming a different prompt group (Please refer to Appendix~\ref{app:prompts} Table~\ref{tab:different_prompts} for details).
We also experiment with mixing the generated data by different prompt groups (mixed).
The results are shown in Table~\ref{tab:prompt_study}.
Overall, the model performance under different prompts is quite close, except on RTE whose test set is very small, potentially resulting in the higher variance. 
% The original prompt group works well across all three tasks, which might imply that there exist ``optimal'' prompts that almost consistently work better than others on different tasks---likely related to the pretraining data distribution and the generator PLM.
In this work, we manually choose simple prompts that make intuitive sense, and we leave the automatic searching of optimal prompts as future work.

% \subsection{Using Different PLMs}
% \label{sec:plms}
% % \input{Tables/plm-ablation}
% The final performance is also relevant to the choice of PLMs as the generator/classifier.
% Apart from the default PLM choice, we report the results of using GPT-2$_{\text{XLarge}}$ ($1.54$B parameters)~\cite{radford2019language} as the generator and RoBERTa$_{\text{Large}}$ ($356$M parameters)~\cite{liu2019roberta} as the classifier in Table~\ref{tab:plm_ablation} with everything else unchanged.
% When using GPT-2, we change the prompt used for SST-2 to ``The film is bad/terrible/awful.'' for the negative label and ``The film is good/great/excellent.'' for the positive label, since the original prompts used for SST-2 in Table~\ref{tab:prompts} are a part of the control codes of CTRL and cannot be effectively leveraged by GPT-2.
% Overall, both CTRL and GPT-2 are able to generate quality training data for good fine-tuned classifier performance; CTRL consistently yields better results than GPT-2 regardless of the choice of the classifier PLM, probably because CTRL is pretrained with control codes which provide explicit guidance for generating texts of certain domains and attributes.
% We also observe that the generated text quality is strongly correlated to the generator's model size---using a smaller version of GPT-2 (\eg, with $117$M parameters) results in significantly less coherent texts and can hardly serve as training data.
% An interesting future direction is to try larger generator PLMs (\eg, GPT-3) which may create training data of better quality.

\subsection{Results with Different Amount of Generated Data}
% \input{Figs/num_train_data}

% \begin{figure}\TopFloatBoxes

% \begin{floatrow}

% \ffigbox[\FBwidth]{
% \begin{subfigure}[t]{0.3\textwidth}
% \centering
% \includegraphics[width=\textwidth]{Figs/MNLI-m.pdf}
% \caption{MNLI-m}
% \end{subfigure}%
% ~
% \begin{subfigure}[t]{0.3\textwidth}
% \centering
% % 	\label{fig:efficency_mm}
% 	\includegraphics[width=\textwidth]{Figs/SST-2.pdf}
% 	\caption{MNLI-mm}
% \end{subfigure}
% \caption{Results with different amount of generated training data used. Dots and error bars are the average performance and the standard deviation over $5$ seeds, respectively.\label{fig:num_train_data} 
% }
% }

% \hspace{0.1cm}

% \ffigbox[\FBwidth]{
% \caption{The performance of \model{}$_\text{Base}$ when pretrained with different crop fractions. The $x$-axis is the fraction of $X^\text{orig}$ being kept (no cropping is $100\%$).}
% \label{fig:fewshot_num_data}
% \vspace{-0.2em}
% \includegraphics[width=0.31\textwidth]{Figs/fewshot-MNLI-m.pdf}
% }

% \end{floatrow}
% % \vspace{-1em}
% \end{figure}

\begin{figure}[t]
\begin{minipage}{0.64\textwidth}
\begin{subfigure}[t]{0.48\textwidth}
\centering
\includegraphics[width=\textwidth]{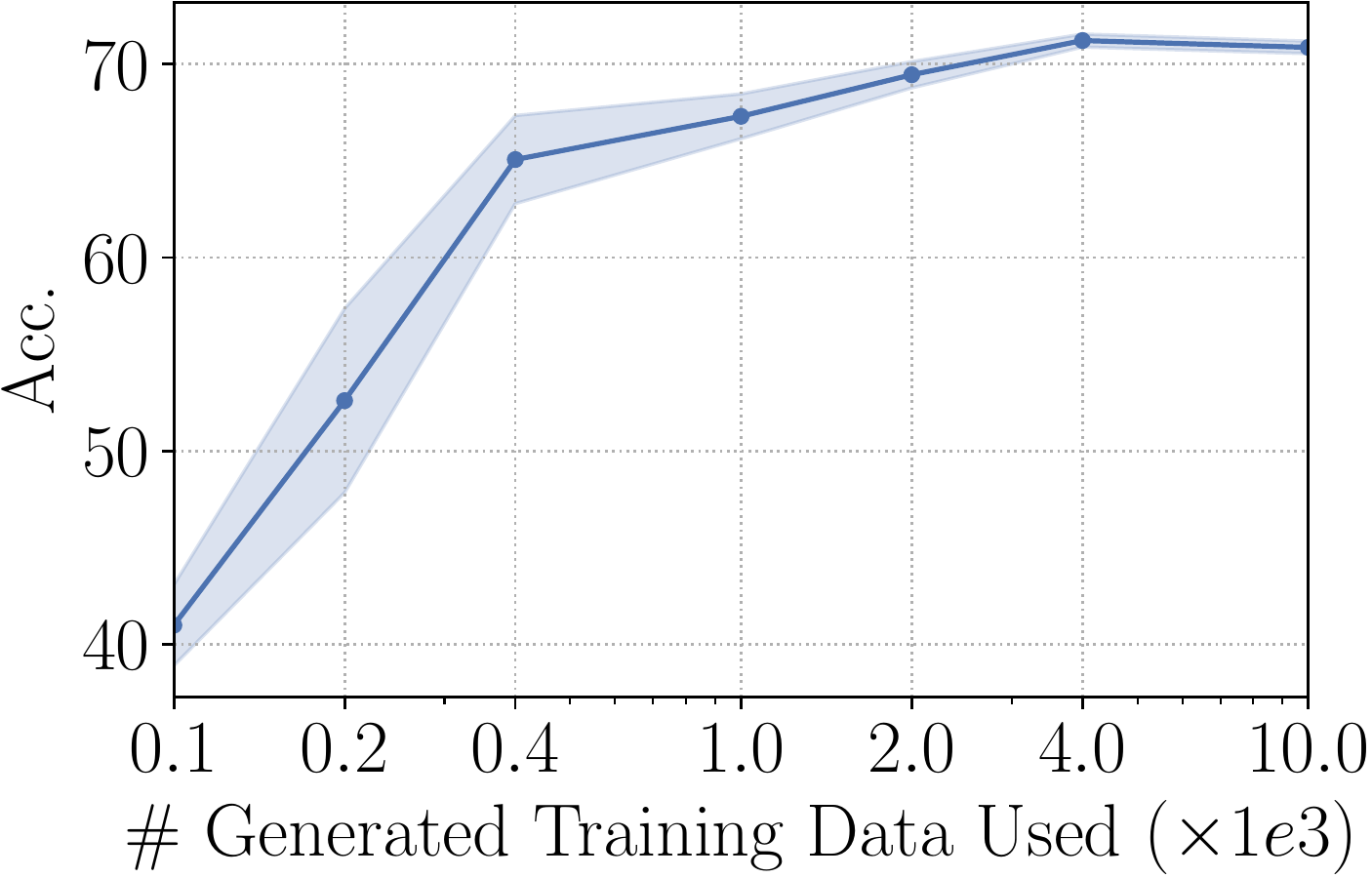}
% \vspace{-0.5em}
\caption{MNLI-m.}
\end{subfigure}%
~
\begin{subfigure}[t]{0.48\textwidth}
\centering
% 	\label{fig:efficency_mm}
\includegraphics[width=\textwidth]{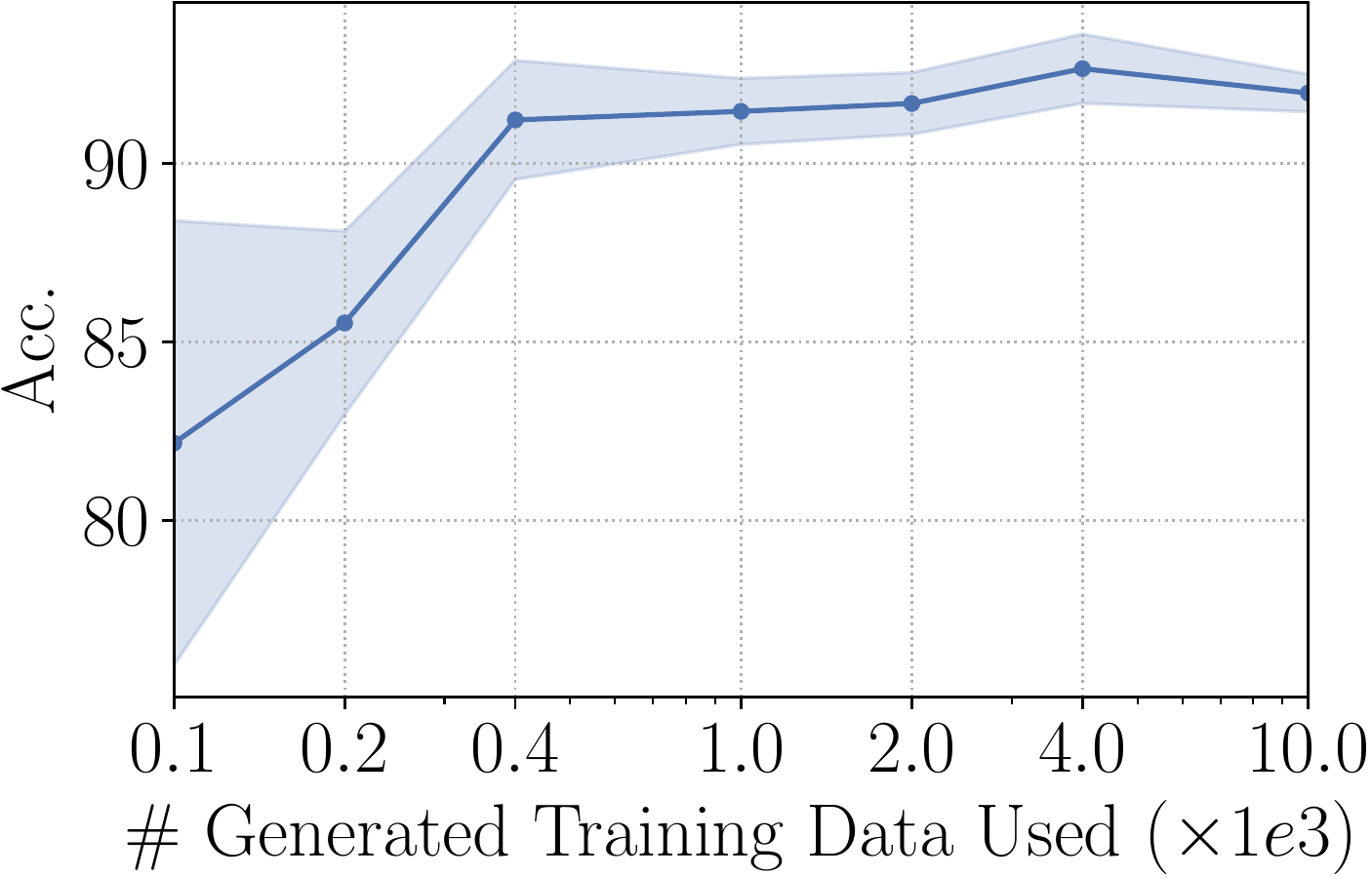}
% \vspace{-0.5em}
\caption{SST-2.}
\end{subfigure}
\vspace{-0.1em}
\caption{Classifier accuracy fine-tuned on different amount of generated training data (after data selection). Dots and error bars are the average performance and the standard deviation over $5$ seeds, respectively. \label{fig:num_train_data} 
}
\end{minipage}
\hfill
\begin{minipage}{0.32\textwidth}
% \vspace{-0.2em}
\centering
\includegraphics[width=0.95\textwidth]{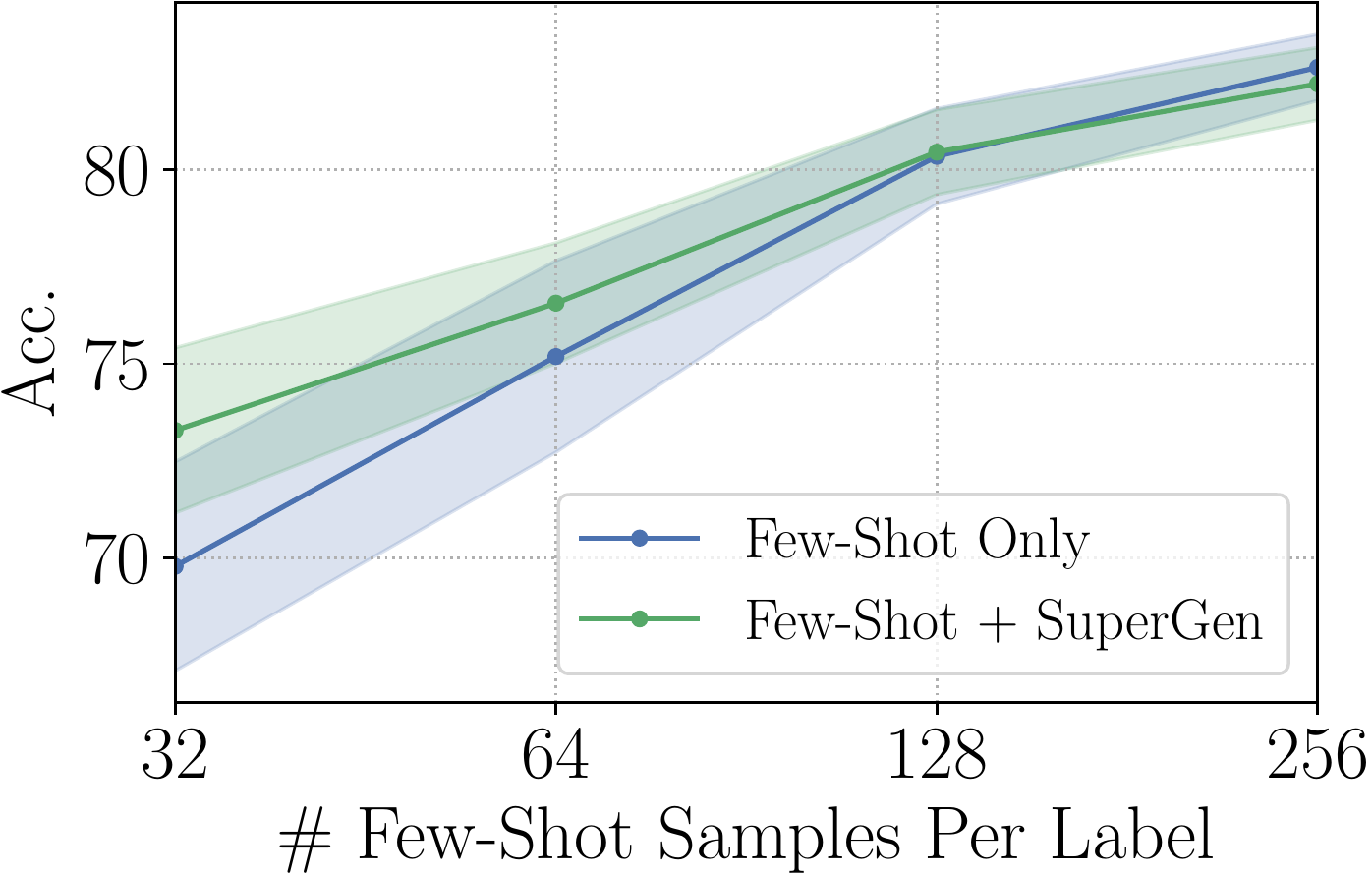}
\vspace{+0.5em}
\caption{Classifier accuracy on MNLI-m fine-tuned on the few-shot samples only vs. on the few-shot and \method generated set with varying few-shot set sizes.}
\label{fig:fewshot_num_data}
\end{minipage}
% \hspace{0.1cm}
\vspace{-1.5em}
\end{figure}

With training data automatically created by the generator, we can have a virtually infinite amount of training samples.
We show the results of using different amount of generated data (after quality data selection) for fine-tuning the classifier $C_{\phi}$ in Fig.~\ref{fig:num_train_data} on MNLI-m and SST-2.
When the number of training data is small (\eg, $100$), the fine-tuning variance is high, resulting in the similar instability issue with few-shot settings.
With more generated data used, both average performance and training stability improve, yielding comparable results (with smaller variance) to fine-tuning using few-shot task-specific data.
However, when too many generated data (\eg, $10,000$) are used, the classifier's performance slightly drops, probably due to increased label noise---recall that the training data are selected based on the ranking score in Eq.~\eqref{eq:sorting_score}, so using more data results in the inclusion of more lower-ranking texts in the training set and reduced data quality.
One way to address this issue is to use a fixed selection ratio and increase the total number of generated texts to obtain a larger number of high-quality training data.
However, this comes at a greater computation cost in the generation step.
An important future direction is thus to develop better data selection strategies.

\subsection{Using \method in Few-Shot Settings}
\label{sec:fewshot}
We present a simple extension of \method to few-shot settings and show that the generated data of \method may also improve the few-shot performance. 
When few-shot samples are available, we first fine-tune the classifier on the few-shot training set (standard prompt-based fine-tuning without regularization), and then continue fine-tuning the classifier on the generated data by \method as described in Section~\ref{sec:fine-tune}.
This allows the classifier to effectively leverage the knowledge from the few-shot training set to filter out noisy samples in the generated data, as temporal ensembling regularizes the classifier to remember the predictions learned previously and only keeps samples on which the model predictions agree with the label.
We show the benefits of incorporating generated data for different few-shot sample sizes on MNLI in Fig.~\ref{fig:fewshot_num_data} (we use half labeled samples for classifier training and half for development): When the few-shot training and validation sets are rather small ($32-64$ samples per label in total), fine-tuning the classifier on the \method generated set further (after fine-tuning on the few-shot samples) brings notable performance improvements.
However, such benefits diminish with more few-shot training samples: The generated data fail to improve the few-shot performance when there are $128$ samples per label, and even worsen the classifier performance with $256$ samples per label.
% We note that few-shot samples are not used in the training data generation stage, and we expect the results to be even better if they are leveraged to generate training data closer to the task-specific distribution.
This is probably because our synthetic data generation process is zero-shot and does not leverage any few-shot samples; the resulting generated samples may not be of high enough quality to boost the few-shot performance when there are relatively abundant annotated samples.
Possible ways to use few-shot samples for generation include using them as demonstrations~\cite{Brown2020LanguageMA}, for creating augmentations~\cite{lee2021neural} and for tuning the generators. 
We leave the explorations of generating higher quality data by leveraging few-shot samples for future work.

\begin{table}[t]
\centering
\begin{minipage}{0.48\textwidth}
\caption{
Comparisons with using CTRL for zero-shot prompting and for knowledge distillation. $^\dagger$: The entire training set is used as unlabeled data. 
}
\vspace{0.5em}
\small 
% \resizebox{0.45\columnwidth}{!}{
\begin{tabular}{lll}
\toprule
\textbf{Method} & \textbf{MNLI-(m/mm)} & \textbf{SST-2} \\
\midrule
SuperGen & \fullres{\textbf{72.3}}{0.5}/\fullres{\textbf{73.8}}{0.5} & \fullres{\textbf{92.8}}{0.6} \\
CTRL Prompting & \fullres{38.5}{0.0}/\fullres{39.2}{0.0} & \fullres{72.5}{0.0} \\
Knowledge Distill$^\dagger$ & \fullres{40.8}{0.5}/\fullres{41.5}{0.6} & \fullres{73.6}{0.8} \\
\bottomrule
\end{tabular}
% }
% \vspace{-1em}
\label{tab:kd_res}
\end{minipage}
\hfill
\begin{minipage}{0.48\textwidth}
\caption{
Results with different generator/classifier PLMs.
}
\vspace{0.6em}
\small 
\begin{tabular}{l*{2}{c}}
\toprule
{\textbf{PLMs} ($G_{\theta}$/$C_{\phi}$)} & \textbf{MNLI-(m/mm)} & \textbf{SST-2} \\ 
\midrule
CTRL/COCO-LM & \fullres{\textbf{72.3}}{0.5}/\fullres{\textbf{73.8}}{0.5} &  \fullres{92.8}{0.6} \\
CTRL/RoBERTa & \fullres{69.0}{0.8}/\fullres{70.6}{0.9} &  \fullres{\textbf{93.0}}{1.5} \\
GPT-2/COCO-LM & \fullres{69.5}{1.2}/\fullres{71.3}{1.3} & \fullres{88.2}{1.8} \\
GPT-2/RoBERTa & \fullres{68.3}{0.9}/\fullres{69.7}{0.7} & \fullres{88.6}{0.8} \\
\bottomrule
\end{tabular}
\label{tab:plm_ablation}
\end{minipage}
\vspace{-0.5em}
\end{table}
\subsection{Using Generators for Knowledge Distillation}
Apart from using unidirectional PLMs $G_{\theta}$ for training data generation, one could also directly apply them to unlabeled data formulated as prompts to obtain zero-shot predictions (\ie, prompting~\cite{Brown2020LanguageMA,gao2021making}), which can then be used as soft labels to train the classifier $C_{\phi}$.
In Table~\ref{tab:kd_res}, we show (1) the zero-shot prediction accuracy of CTRL (the best out of three different prompts, details in Appendix~\ref{app:kd_details}) and (2) the classifier performance trained from CTRL's predictions on the entire unlabeled training set as soft labels (\ie, knowledge distillation).
Similar to the observations in previous studies~\cite{Brown2020LanguageMA,Wei2022FinetunedLM,Zhao2021CalibrateBU}, the zero-shot predictions of unidirectional PLMs are quite inaccurate and directly using them as soft labels to train classifiers does not yield good results.
% Notably, the 
We hypothesize that the advantages of using unidirectional PLMs for training data generation over using them for zero-shot predictions are twofold: 
(1) Better flexibility in prompt formats. When unidirectional PLMs are used for zero-shot predictions, the prompts have to be designed so that the label word is the last token in the sequence to be predicted, as unidirectional PLMs cannot attend to subsequent tokens.
Such constraints may result in the prompt being dissimilar to the pretraining data distribution and worsen the prediction quality of the PLMs.
On the contrary, using unidirectional PLMs for generation is not subject to any prompt format constraints.
(2) More direct uses of PLMs' language modeling ability. 
Using unidirectional PLMs for training data generation \emph{directly} leverages the PLMs' output token probability.
Applying PLMs for zero-shot prediction, however, requires an additional step to convert token predictions to label predictions (\ie, the verbalizer~\cite{Schick2021ExploitingCF}), and such a mapping process usually necessitates manual curation and can hardly be optimal~\cite{gao2021making} especially without abundant task-specific data.

\subsection{Using Different PLMs}
\label{sec:plms}
The final performance of \method is relevant to the choice of PLMs as the generator/classifier.
Apart from the default PLM choice, we report the results of using GPT-2$_{\text{XLarge}}$ ($1.54$B parameters)~\cite{radford2019language} as the generator and RoBERTa$_{\text{Large}}$ ($356$M parameters)~\cite{liu2019roberta} as the classifier in Table~\ref{tab:plm_ablation} with everything else unchanged.
When using GPT-2, we change the prompt used for SST-2 to ``The film is bad/terrible/awful.'' for the negative label and ``The film is good/great/excellent.'' for the positive label, since the original prompts used for SST-2 in Table~\ref{tab:prompts} are a part of the control codes of CTRL and cannot be effectively leveraged by GPT-2.
Overall, both CTRL and GPT-2 are able to generate quality training data for good fine-tuned classifier performance; CTRL consistently yields better results than GPT-2 regardless of the choice of the classifier PLM, probably because CTRL is pretrained with control codes which provide explicit guidance for generating texts of certain domains and attributes.
We also observe that the generated text quality is strongly correlated to the generator's model size---using a smaller version of GPT-2 (\eg, with $117$M parameters) results in significantly less coherent texts and can hardly serve as training data.
An interesting future direction is to try larger generator PLMs (\eg, GPT-3) which may create training data of better quality.

\subsection{Case Studies}
\begin{table}[thb]
\renewcommand\arraystretch{1.2}
\caption{
Example generated texts for SST-2, MNLI and QQP. \textit{Sampled sequences} from pretraining corpus ($\bs{x}^s$) are italicized; \underline{generated sequences} ($\bs{x}^g$) are underlined; \textbf{prompts} ($\bs{w}^y$) are in bold.
}
\vspace{0.5em}
\centering
\small 
\resizebox{\textwidth}{!}{
\begin{tabular}{lll}
\toprule
\textbf{Task} & \textbf{Label} & \textbf{Generated Text} \\
\midrule
\multirow{2}{*}{\textbf{SST-2}} & positive & \textbf{Rating: 5.0} \underline{The film is a great example of the kind of movie that you can watch over and over.} \\
& negative & \textbf{Rating: 1.0} \underline{The film was a total waste of time. I would not recommend this movie to anyone.} \\
\midrule
\multirow{6}{*}{\textbf{MNLI}} & \multirow{2}{*}{entailment} & \textit{The construction of the station began in 2016, when the opening date was estimated to mid-2020.} \\
& & \textbf{In other words}, \underline{the station was to open in 2020.} \\
& \multirow{2}{*}{neutral} & \textit{The construction of the station began in 2016, when the opening date was estimated to mid-2020.} \\
& & \textbf{Furthermore}, \underline{it is expected that a new bus terminal will be built at this station.} \\
& \multirow{2}{*}{contradiction} & \textbf{There is a rumor that} \textit{The construction of the station began in 2016, when the opening date was estimated to mid-2020.} \\
& & \textbf{However, the truth is:} \underline{The construction started in 2017, and the official opening date was set for March 31, 2018.} \\
\midrule
\multirow{2}{*}{\textbf{QQP}} & equivalent & \textit{What are the most wear resistant steels?} \textbf{In other words,} \underline{what are the most durable steels?} \\
& not equivalent & \textit{What are the most wear resistant steels?} \textbf{Furthermore,} \underline{what is the best way to clean them?} \\
\bottomrule
\end{tabular}
}
\vspace{-1em}
\label{tab:case_studies}
\end{table}

We present concrete examples of generated texts guided by prompts of different labels in Table~\ref{tab:case_studies}.
The generated sequences are not only coherent, but also pertain to the corresponding labels.
For easier tasks like SST-2, the generated texts almost always correctly reflect the desired sentiment polarity specified by the prompt. 
For more difficult tasks like MNLI, sometimes the generated texts are not of the correct label (Appendix~\ref{app:neg_res} Table~\ref{tab:negative_gen} shows some negative results).
The existence of such label noise motivates our use of the regularization techniques in the fine-tuning stage.
In the future, it will be interesting to develop new methods to better control text generation towards the desired label.

\section{Discussions and Conclusions}
\paragraph{Ethical Considerations.}
While PLMs have demonstrated remarkable text generation and understanding capability, they can come with potential risks or harms~\cite{Bender2020ClimbingTN,Bender2021OnTD,Brown2020LanguageMA} such as generating misinformation~\cite{Pagnoni2021UnderstandingFI} or amplifying harmful biases~\cite{Prabhumoye2018StyleTT}. The focus of our work is on utilizing existing PLMs to generate training data for NLU tasks instead of developing new PLMs or generation methods.
Therefore, our method can be used in company with any bias reduction and correction techniques~\cite{Gehman2020RealToxicityPromptsEN,Ma2020PowerTransformerUC} to mitigate the risks of PLMs.

\paragraph{Limitations.}
One inherent limitation with zero-shot learning is the lack of access to task-specific samples for hyperparameter tuning, whereas the performance of neural networks is usually heavily dependent on the choice of hyperparameters even when the training algorithm and training set are fixed~\cite{Perez2021TrueFL}.
% Therefore, under the fully zero-shot learning setting, it is almost impossible to use the optimal hyperparameters for model training.
Also, without access to any labeled data, the generated training data quality may not be high enough to achieve good performance on challenging tasks, especially when the task distribution is significantly different from the pretraining data distribution (\eg, the ``linguistically incorrect'' label of CoLA requires generating sequences with grammar mistakes -- a different distribution from the one used to train PLMs).
A promising direction to address the above limitations is extending \method to few-shot settings (\eg, the setting studied in Section~\ref{sec:fewshot}) and leveraging a small amount of labeled data for generating better quality data and for hyperparameter tuning.

\paragraph{Conclusions.}
We propose \method, an automatic supervision generation approach for zero-shot learning of NLU tasks. 
By providing label-descriptive prompts as guidance to a unidirectional PLM, training data can be automatically created for fine-tuning a bidirectional PLM.
Our framework differs from previous transfer-learning-based zero-shot methods in that \method does not rely on cross-task annotations and eliminates the task difference in training and inference.
We show that several strategies are important for effective and stable fine-tuning on generated data, including quality training data selection, label smoothing and temporal ensembling.
\method achieves strong performance on seven classification tasks of the GLUE benchmark, even yielding comparable or better results than sophisticated few-shot learning methods and offering better stability.
There is large room for future work, including but not limited to: Extension to few-shot learning settings, exploring larger generator models~\cite{Kim2021WhatCC,wang2021towards}, better fine-tuning techniques to leverage generated data and better strategies for selecting quality training data.

\section*{Acknowledgments}
Research was supported in part by US DARPA KAIROS Program No.\ FA8750-19-2-1004 and INCAS Program No.\ HR001121C0165, National Science Foundation IIS-19-56151, IIS-17-41317, and IIS 17-04532, and the Molecule Maker Lab Institute: An AI Research Institutes program supported by NSF under Award No.\ 2019897, and the Institute for Geospatial Understanding through an Integrative Discovery Environment (I-GUIDE) by NSF under Award No.\ 2118329. Any opinions, findings, and conclusions or recommendations expressed herein are those of the authors and do not necessarily represent the views, either expressed or implied, of DARPA or the U.S. Government.
Yu Meng is supported by the Google PhD Fellowship.
We thank anonymous reviewers for valuable and insightful feedback.

%%%%%%%%%%%%%%%%%%%%%%%%%%%%%%%%%%%%%%%%%%%%%%%%%%%%%%%%%%%
% \bibliographystyle{unsrtnat}
\bibliographystyle{plainnat}
\bibliography{ref}

%%%%%%%%%%%%%%%%%%%%%%%%%%%%%%%%%%%%%%%%%%%%%%%%%%%%%%%%%%%

% \newpage

\section*{Checklist}

% %%% BEGIN INSTRUCTIONS %%%
% The checklist follows the references.  Please
% read the checklist guidelines carefully for information on how to answer these
% questions.  For each question, change the default \answerTODO{} to \answerYes{},
% \answerNo{}, or \answerNA{}.  You are strongly encouraged to include a {\bf
% justification to your answer}, either by referencing the appropriate section of
% your paper or providing a brief inline description.  For example:
% \begin{itemize}
%   \item Did you include the license to the code and datasets? \answerYes{See Section~\ref{gen_inst}.}
%   \item Did you include the license to the code and datasets? \answerNo{The code and the data are proprietary.}
%   \item Did you include the license to the code and datasets? \answerNA{}
% \end{itemize}
% Please do not modify the questions and only use the provided macros for your
% answers.  Note that the Checklist section does not count towards the page
% limit.  In your paper, please delete this instructions block and only keep the
% Checklist section heading above along with the questions/answers below.
% %%% END INSTRUCTIONS %%%

\begin{enumerate}

\item For all authors...
\begin{enumerate}
  \item Do the main claims made in the abstract and introduction accurately reflect the paper's contributions and scope?
    \answerYes{}
  \item Did you describe the limitations of your work?
    \answerYes{}
  \item Did you discuss any potential negative societal impacts of your work?
    \answerYes{}
  \item Have you read the ethics review guidelines and ensured that your paper conforms to them?
    \answerYes{}
\end{enumerate}

\item If you are including theoretical results...
\begin{enumerate}
  \item Did you state the full set of assumptions of all theoretical results?
    \answerNA{}
  \item Did you include complete proofs of all theoretical results?
    \answerNA{}
\end{enumerate}

\item If you ran experiments...
\begin{enumerate}
  \item Did you include the code, data, and instructions needed to reproduce the main experimental results (either in the supplemental material or as a URL)?
    \answerYes{}
  \item Did you specify all the training details (e.g., data splits, hyperparameters, how they were chosen)?
    \answerYes{}
        \item Did you report error bars (e.g., with respect to the random seed after running experiments multiple times)?
    \answerYes{}
        \item Did you include the total amount of compute and the type of resources used (e.g., type of GPUs, internal cluster, or cloud provider)?
    \answerYes{}
\end{enumerate}

\item If you are using existing assets (e.g., code, data, models) or curating/releasing new assets...
\begin{enumerate}
  \item If your work uses existing assets, did you cite the creators?
    \answerYes{}
  \item Did you mention the license of the assets?
    \answerYes{}
  \item Did you include any new assets either in the supplemental material or as a URL?
    \answerNA{}
  \item Did you discuss whether and how consent was obtained from people whose data you're using/curating?
    \answerNA{}
  \item Did you discuss whether the data you are using/curating contains personally identifiable information or offensive content?
    \answerNA{}
\end{enumerate}

\item If you used crowdsourcing or conducted research with human subjects...
\begin{enumerate}
  \item Did you include the full text of instructions given to participants and screenshots, if applicable?
    \answerNA{}
  \item Did you describe any potential participant risks, with links to Institutional Review Board (IRB) approvals, if applicable?
    \answerNA{}
  \item Did you include the estimated hourly wage paid to participants and the total amount spent on participant compensation?
    \answerNA{}
\end{enumerate}

\end{enumerate}

% \newpage
\appendix

\section{Details of Prompts Used for Different Tasks}
\label{app:prompts}
\begin{table}[h]
\caption{
Extensions of Table~\ref{tab:prompts} with more details of prompts used to generate class-conditioned texts for different GLUE tasks.
SST-2 and CoLA are single-sequence classification tasks and the rest are sequence-pair classification tasks. 
Generation for CoLA does not use prompts but by varying sampling temperatures.
Text generation with CTRL~\cite{Keskar2019CTRLAC} requires starting with control codes, and we use the ones that correspond to the pretraining corpus where the first sequence is sampled: For MNLI, RTE and MRPC, the first sequence is sampled from Wikipedia; for QNLI and QQP, the first sequence is sampled from OpenWebText~\cite{Gokaslan2019OpenWeb}.
$\bs{x}^s$ denotes a sequence randomly sampled from the pretraining corpus; $\bs{x}^g$ denotes the sequence to be generated by $G_{\theta}$; $\dots$ denotes skipping at least one sequence. 
% The ``not entailment'' label of MRPC and ``not equivalent'' label of RTE use two prompts split by ``//'' (essentially combining the prompts used for ``neutral'' and ``contradiction'' labels of the MNLI task).
The prompts used for SST-2 are part of the CTRL~\cite{Keskar2019CTRLAC} codes.
}
\vspace{0.5em}
\centering
\small 
\resizebox{\columnwidth}{!}{
\begin{tabular}{lllll}
\toprule
\textbf{Task} & \textbf{Task Type} & \textbf{Control Code} & \textbf{Label} & \textbf{Prompt} \\
\midrule
\multirow{2}{*}{\textbf{SST-2}} & \multirow{2}{*}{single-sequence} & \multirow{2}{*}{Reviews} & positive & Rating: 5.0 $\bs{x}^g$ \\
& & & negative & Rating: 1.0 $\bs{x}^g$ \\
\midrule
\multirow{2}{*}{\textbf{CoLA}} & \multirow{2}{*}{single-sequence} & \multirow{2}{*}{Links} & grammatical & $\bs{x}^g$ \\
& & & not grammatical & $\bs{x}^g$ \\
\midrule
\multirow{3}{*}{\textbf{MNLI}} & \multirow{3}{*}{sequence-pair} & \multirow{3}{*}{Wikipedia} & entailment & $\bs{x}^s$. In other words, $\bs{x}^g$ \\
& & & neutral & $\bs{x}^s$. Furthermore, $\bs{x}^g$ \\
& & & contradiction & There is a rumor that $\bs{x}^s$. However, the truth is: $\bs{x}^g$ \\
\midrule
\multirow{2}{*}{\textbf{QNLI}} & \multirow{2}{*}{sequence-pair} & \multirow{2}{*}{Links} & entailment & $\bs{x}^s$? $\bs{x}^g$ \\
& & & not entailment & $\bs{x}^s$? \dots $\bs{x}^g$ \\
\midrule
\multirow{2}{*}{\textbf{RTE}} & \multirow{2}{*}{sequence-pair} & \multirow{2}{*}{Wikipedia} & entailment & $\bs{x}^s$. In other words, $\bs{x}^g$ \\
& & & not entailment & $\bs{x}^s$. Furthermore, $\bs{x}^g$ \\
% // There is a rumor that $\bs{x}^s$. However, the truth is: $\bs{x}^g$ \\
\midrule
\multirow{2}{*}{\textbf{MRPC}} & \multirow{2}{*}{sequence-pair} & \multirow{2}{*}{Wikipedia} & equivalent & $\bs{x}^s$. In other words, $\bs{x}^g$ \\
& & & not equivalent & $\bs{x}^s$. Furthermore, $\bs{x}^g$ \\
% // There is a rumor that $\bs{x}^s$. However, the truth is: $\bs{x}^g$ \\
\midrule
\multirow{2}{*}{\textbf{QQP}} & \multirow{2}{*}{sequence-pair} & \multirow{2}{*}{Links} & equivalent & $\bs{x}^s$? In other words, $\bs{x}^g$ \\
& & & not equivalent & $\bs{x}^s$? Furthermore, $\bs{x}^g$ \\
\bottomrule
\end{tabular}
}
\vspace{-1.5em}
\label{tab:full_prompts}
\end{table}

\begin{table}[h]
\renewcommand\arraystretch{1.2}
\caption{
Different prompt groups used in the experiments of Section~\ref{sec:diff_prompts}. 
We replace the original prompt for each label with an alternative one and keep other prompts unchanged when forming a different prompt group.
}
\vspace{0.5em}
\centering
\small 
\resizebox{\columnwidth}{!}{
\begin{tabular}{llll}
\toprule
\textbf{Task} & \textbf{Label} & \textbf{Original} & \textbf{Alternative} \\
\midrule
\multirow{2}{*}{\textbf{SST-2}} & positive  & Rating: 5.0 $\bs{x}^g$ & Rating: 4.0 $\bs{x}^g$ \\
& negative & Rating: 1.0 $\bs{x}^g$ & Rating: 2.0 $\bs{x}^g$ \\
\midrule
\multirow{3}{*}{\textbf{MNLI}} & entailment  & $\bs{x}^s$. In other words, $\bs{x}^g$ & $\bs{x}^s$. To put it another way, $\bs{x}^g$ \\
& neutral & $\bs{x}^s$. Furthermore, $\bs{x}^g$ & $\bs{x}^s$. In addition, $\bs{x}^g$ \\
& contradiction & There is a rumor that $\bs{x}^s$. However, the truth is: $\bs{x}^g$ & People believe that $\bs{x}^s$. However, the truth is: $\bs{x}^g$ \\
\midrule
\multirow{2}{*}{\textbf{QNLI}} & entailment  & $\bs{x}^s$? $\bs{x}^g$ & Question: $\bs{x}^s$? Answer: $\bs{x}^g$ \\
& not entailment & $\bs{x}^s$? \dots $\bs{x}^g$ & Question: $\bs{x}^s$? Answer: \dots $\bs{x}^g$ \\
\midrule
\multirow{2}{*}{\textbf{RTE}} & entailment  & $\bs{x}^s$. In other words, $\bs{x}^g$ & $\bs{x}^s$. To put it another way, $\bs{x}^g$ \\
& not entailment & $\bs{x}^s$. Furthermore, $\bs{x}^g$ & $\bs{x}^s$. In addition, $\bs{x}^g$ \\
\midrule
\multirow{2}{*}{\textbf{MRPC}} & equivalent  & $\bs{x}^s$. In other words, $\bs{x}^g$ & $\bs{x}^s$. To put it another way, $\bs{x}^g$ \\
& not equivalent & $\bs{x}^s$. Furthermore, $\bs{x}^g$ & $\bs{x}^s$. In addition, $\bs{x}^g$ \\
\midrule
\multirow{2}{*}{\textbf{QQP}} & equivalent  & $\bs{x}^s$? In other words, $\bs{x}^g$ & $\bs{x}^s$? To put it another way, $\bs{x}^g$ \\
& not equivalent & $\bs{x}^s$? Furthermore, $\bs{x}^g$ & $\bs{x}^s$? In addition, $\bs{x}^g$ \\
\bottomrule
\end{tabular}
}
\vspace{-1.5em}
\label{tab:different_prompts}
\end{table}

We present more details about the prompts used for different tasks in Table~\ref{tab:full_prompts} which is an extended version of Table~\ref{tab:prompts}.

% RTE and MRPC use two prompts for the ``not entailment'' label and the ``not equivalent'' label, by combining the ``neutral'' and ``contradiction'' prompts from MNLI.
% The two prompts will be used to generate the same number of training samples.
% Therefore, RTE and MRPC essentially share the same set of generated training data with MNLI.

For SST-2, we fix the beginning of the generated sequence $\bs{x}^g$ to be ``The/this film/movie'' to make sure the generated texts are related to movie reviews. 
For CoLA, we start the generated sequence $\bs{x}^g$ with a random stop word.
For QNLI and QQP, the first sequence is always a question, and we require the sampled sequence $\bs{x}^s$ to end with a question mark and begin with one of the following words: ``how", ``what'', ``why'', ``who'', ``which'', ``where'', ``when'', ``whom'', ``whose''.
For QNLI, the generated sequence $\bs{x}^g$ for the ``entailment'' label is the one that immediately follows the sampled sequence $\bs{x}^s$; the generated sequence $\bs{x}^g$ for the ``not entailment'' label is randomly sampled from the paragraph following $\bs{x}^g$ excluding the first sequence that immediately follows $\bs{x}^g$.

We also show the different prompt groups used in the experiments of Section~\ref{sec:diff_prompts} in Table~\ref{tab:different_prompts}.

\section{Hyperparameters and Reproducibility}
\label{app:hyperpara}
\paragraph{Hyperparameters for Generating Training Data.} 

% \begin{wraptable}{r}{8.5cm}
\begin{table}[t]
\caption{
Hyperparameters for generating training data of different tasks.
$\tau$: Temperature during sampling ($\tau = 0$ means using greedy sampling); $\alpha$ and $\beta$: Repetition rewarding/penalizing parameters; $M$: Number of total generated texts per label. The top-$k$ sampling (if $\tau > 0$) uses $k=10$.
}
\vspace{0.5em}
\centering
\small 
% \resizebox{0.5\columnwidth}{!}{
\begin{tabular}{ll*{4}{c}}
\toprule
\textbf{Task} & \textbf{Label} & $\tau$ & $\alpha$ & $\beta$ & $M$ \\
\midrule
\multirow{2}{*}{\textbf{SST-2}} & positive & \multirow{2}{*}{0.2} & - & 1.2 & 25,000 \\
& negative &  & - & 1.2 & 25,000 \\
\midrule
\multirow{2}{*}{\textbf{CoLA}} & grammatical & \multirow{2}{*}{[0.1, 10]} & - & 1.2 & 10,000 \\
& not grammatical & & - & 1.2 & 10,000 \\
\midrule
\multirow{3}{*}{\textbf{MNLI}} & entailment & \multirow{3}{*}{0} & 0.8 & 1.1 & 25,000 \\
& neutral &  & 1.3 & 1.3 & 25,000 \\
& contradiction &  & 1.1 & 1.1 & 25,000 \\
\midrule
\multirow{2}{*}{\textbf{QNLI}} & entailment & \multirow{2}{*}{0} & 0.9 & 1.2 & 25,000 \\
& not entailment &  & 0.9 & 1.2 & 25,000 \\
\midrule
\multirow{2}{*}{\textbf{RTE}} & entailment & \multirow{2}{*}{0} & 0.8 & 1.1 & 30,000 \\
& not entailment & & 1.1 & 1.1 & 30,000 \\
\midrule
\multirow{2}{*}{\textbf{MRPC}} & equivalent & \multirow{2}{*}{0} & 0.8 & 1.1 & 30,000 \\
& not equivalent & & 1.1 & 1.1 & 30,000 \\
\midrule
\multirow{2}{*}{\textbf{QQP}} & equivalent & \multirow{2}{*}{0} & 1.0 & 1.2 & 25,000 \\
& not equivalent & & 1.2 & 1.2 & 25,000 \\
\bottomrule
\end{tabular}
% }
\label{tab:gen_hyperpara}
\vspace{-1em}
\end{table}
% \end{wraptable}
Table~\ref{tab:gen_hyperpara} lists the hyperparameters used in the training data generation stage.
For sequence-pair tasks, we use greedy sampling for better reproducibility.
For labels that require generating entailment, paraphrase, or equivalent sequence pairs, we set $\alpha \le 1$ to encourage word overlapping between the second sequence and the first sequence; otherwise, we set $\alpha = \beta > 1$ to discourage word repetition. 

To construct a training set consisting of $N$ samples per class, we will generate $M$ samples per class, and select training data based on the score $r$ in Eq.~\eqref{eq:sorting_score}: 
For all tasks except CoLA and the ``neutral'' label of MNLI, the top-$N$ ones of each class are selected; for CoLA, the top-$N$ ones are used as the training sample as linguistically acceptable sequences, and the bottom-$N$ ones are as linguistically unacceptable sequences; for the ``neutral'' label of MNLI, we find it better to randomly select $N$ samples from the total $M$ samples instead of using the ranking score, probably because a neutral hypothesis with respect to the premise has a wide range of possibilities (\ie, any hypothesis that is not entailed by or contradicts with the premise will be neutral), and random selection improves the diversity in generated hypotheses of the neutral label.

\paragraph{Hyperparameters for Fine-Tuning.}
\begin{table}[t]
\caption{
Hyperparameters used for fine-tuning on different tasks (they are kept same for all tasks).
Fine-tuning-related hyperparameters (\eg, learning rate, batch size) follow the default values (when the validation set is not available) in Appendix A of \cite{gao2021making}; 
regularization-related hyperparameters follow the default values in label smoothing and temporal ensembling.
$lr$: Learning rate; $bs$: Batch size; $N|\mathcal{Y}|$: Total number of selected generated data (\ie, training set size); $B$: Ensemble prediction update interval; $T$: Number of training steps; $\epsilon$: Label smoothing parameter; $\gamma$: Temporal ensembling momentum parameter; $\delta$: Threshold for filtering out noisy data; $\lambda_{\text{max}}$: Maximum weight (after ramp-up) of temporal ensembling regularization.
}
\vspace{0.5em}
\centering
\small 
% \resizebox{\columnwidth}{!}{
\begin{tabular}{*{9}{c}}
\toprule
$lr$ & $bs$ & $N|\mathcal{Y}|$ & $B$ & $T$ & $\epsilon$ & $\gamma$ & $\delta$ & $\lambda_{\text{max}}$ \\
\midrule
1e-5 & 16 & 6,000 & 100 & 1,125 & 0.15 & 0.8 & 0.8 & 10 \\
\bottomrule
\end{tabular}
% }
\vspace{-1em}
\label{tab:finetune_hyperpara}
\end{table}
Table~\ref{tab:finetune_hyperpara} lists the hyperparameters used in the fine-tuning stage.
We keep them the same across all tasks except CoLA which uses $\delta=0$ because half of the training data for CoLA are intentionally made to be of low quality (\ie, as linguistically unacceptable sequences) and there is no need to filter them out.
We follow \cite{Laine2017TemporalEF} to slowly ramp-up $\lambda$ in Equation~\eqref{eq:train_obj} during the first $10$ ensembles: $\lambda(t) = \lambda_{\text{max}}\exp(-5(1-t/10)^2)$ where $t$ is the number of prediction ensembles performed.
\paragraph{Computation Environment.} 
All experiments are conducted on NVIDIA GeForce RTX 3090 GPUs. 
\method can be run on typical research hardware (\eg, with $>10$GB GPU memory). The generator PLM $G_\theta$ does not need to be trained so a relatively large generator can be used (\eg, a $1.63$B-parameter CTRL model).

\section{GLUE Tasks}
\label{app:glue}
We provide the details of the seven classification tasks included in the GLUE benchmark.

\textbf{MNLI:} Multi-genre Natural Language Inference~\cite{MNLI} aims to predict whether a given premise sentence entails, contradicts or neutral with respect to a given hypothesis sentence. It has two test sets, matched (MNLI-m) and mismatched (MNLI-mm), which correspond to samples from the same sources as the training set and samples that do not resemble the training data, respectively.

\textbf{QQP:} Quora Question Pairs~\cite{QQP} aims to determine whether a pair of questions asked are semantically equivalent.

\textbf{QNLI:} Question Natural Language Inference aims to predict whether a given sentence contains the answer to a given question sentence.

\textbf{SST-2:} Stanford Sentiment Treebank~\cite{SST-2} aims to determine if a movie review has positive or negative sentiment. 

\textbf{CoLA:} Corpus of Linguistic Acceptability~\cite{COLA} aims to determine whether a given sentence is linguistically acceptable or not. 

\textbf{RTE:} Recognizing Textual Entailment~\cite{RTE-5,RTE-1,RTE-3,RTE-2} aims to predict whether a given premise sentence entails a given hypothesis sentence or not.

\textbf{MRPC:} Microsoft Research Paraphrase Corpus~\cite{MRPC} aims to predict whether two sentences are semantically equivalent or not.

\begin{table}[t]
\caption{
Different prompts used on MNLI for CTRL zero-shot prompting and knowledge distillation baselines. $\bs{x}_1$ and $\bs{x}_2$ denote the first and second input sequence, respectively.
}
\vspace{1em}
\centering
\small 
% \resizebox{\columnwidth}{!}{
\begin{tabular}{llll}
\toprule
\textbf{Prompt} & \textbf{Template} & \textbf{Label name}\\
\midrule
\multirow{3}{*}{$\#1$} & Sentence 1: $\bs{x}_1$ Sentence 2: $\bs{x}_2$ & entailment: Yes \\
& Does Sentence 1 entail Sentence 2? & neutral: Maybe \\
& The answer is: &  contradiction: No \\
\midrule
\multirow{3}{*}{$\#2$} & Premise: $\bs{x}_1$ Hypothesis: $\bs{x}_2$ & entailment: Yes \\
& Does the premise entail the hypothesis? & neutral: Maybe \\
& Options: Yes. No. Maybe. The answer is: &  contradiction: No \\
\midrule
\multirow{3}{*}{$\#3$} & Premise: $\bs{x}_1$ Hypothesis: $\bs{x}_2$ & entailment: Entailment \\
& What is the relation between the premise and the hypothesis? & neutral: Neutral \\
& Options: Entailment. Neutral. Contradiction. The answer is: & contradiction: Contradiction \\
\bottomrule
\end{tabular}
% }
\vspace{-1em}
\label{tab:kd_prompts_mnli}
\end{table}

\begin{table}[t]
\caption{
Different prompts used on SST-2 for CTRL zero-shot prompting and knowledge distillation baselines. $\bs{x}$ denotes the input sequence.
}
\vspace{1em}
\centering
\small 
% \resizebox{\columnwidth}{!}{
\begin{tabular}{llll}
\toprule
\textbf{Prompt} & \textbf{Template} & \textbf{Label name}\\
\midrule
$\#1$ & $\bs{x}$ This is & positive: good; negative: bad \\
\midrule
$\#2$ & $\bs{x}$ It was & positive: good; negative: bad \\
\midrule
$\#3$ & Review: $\bs{x}$ Sentiment: & positive: Positive; negative: Negative \\
\bottomrule
\end{tabular}
% }
\label{tab:kd_prompts_sst}
\vspace{-1em}
\end{table}
\section{Knowledge Distillation Baseline Details}
\label{app:kd_details}

We show the concrete prompts used for the knowledge distillation baseline in Tables~\ref{tab:kd_prompts_mnli} and \ref{tab:kd_prompts_sst} on MNLI and SST-2, respectively. We use the best prompt (prompt $\#1$ in both tables) out of the three according to the zero-shot test set prediction accuracy for generating soft labels to train the classification model (\ie, knowledge distillation). The classifier is trained with Kullback–Leibler (KL) divergence as the objective to approximate the soft labels generated by CTRL on the entire training set.

\section{Negative Generation Results}
\label{app:neg_res}
\begin{table}[thb]
\renewcommand\arraystretch{1.2}
\caption{
Negative example generated texts for MNLI that do not pertain to the desired label. \textit{Sampled sequences} from pretraining corpus ($\bs{x}^s$) are italicized; \underline{generated sequences} ($\bs{x}^g$) are underlined; \textbf{prompts} ($\bs{w}^y$) are in bold. In example $\#1$, the generated text corresponding to the ``entailment'' label actually pertains to the ``neutral'' label. In example $\#2$, the generated text corresponding to the ``contradiction'' label actually pertains to the ``neutral'' label. In example $\#3$, the generated text corresponding to the ``neutral'' label actually pertains to the ``contradition'' label.
}
\vspace{1em}
\centering
\small 
\resizebox{\textwidth}{!}{
\begin{tabular}{lll}
\toprule
\textbf{Example} & \textbf{Label} & \textbf{Generated Text} \\
\midrule
\multirow{6}{*}{$\#1$} & \multirow{2}{*}{entailment} & \textit{Directed by Daniel J. Sullivan, the cast features Kelly AuCoin, Annie Parisse, Lee Tergesen and Alex Wolff.} \\
& & \textbf{In other words}, \underline{the cast features a lot of people who are not in the original Broadway cast.} \\
& \multirow{2}{*}{neutral} & \textit{Directed by Daniel J. Sullivan, the cast features Kelly AuCoin, Annie Parisse, Lee Tergesen and Alex Wolff.} \\
& & \textbf{Furthermore}, \underline{it was announced that a sequel to ``The Last Airbender'' will be released in 2019.} \\
& \multirow{2}{*}{contradiction} & \textbf{There is a rumor that} \textit{Directed by Daniel J. Sullivan, the cast features Kelly AuCoin, Annie Parisse, Lee Tergesen and Alex Wolff.} \\
& & \textbf{However, the truth is:} \underline{The cast includes only two actors from the original Broadway production.} \\
\midrule
\multirow{6}{*}{$\#2$} & \multirow{2}{*}{entailment} & \textit{As a result, minor problems begin to look serious and rare events begin to seem common.} \\
& & \textbf{In other words}, \underline{a minor problem can begin to look serious and rare events seem common.} \\
& \multirow{2}{*}{neutral} & \textit{As a result, minor problems begin to look serious and rare events begin to seem common.} \\
& & \textbf{Furthermore}, \underline{the number of people who are affected by these changes is increasing.} \\
& \multirow{2}{*}{contradiction} & \textbf{There is a rumor that} \textit{As a result, minor problems begin to look serious and rare events begin to seem common.} \\
& & \textbf{However, the truth is:} \underline{The only thing that can cause such an effect is a major change in the weather.} \\
\midrule
\multirow{6}{*}{$\#3$} & \multirow{2}{*}{entailment} & \textit{The character was announced on 20 September 2019, but casting wasn't revealed until 10 December 2019.} \\
& & \textbf{In other words}, \underline{the casting wasn't revealed until 10 December 2019.} \\
& \multirow{2}{*}{neutral} & \textit{The character was announced on 20 September 2019, but casting wasn't revealed until 10 December 2019.} \\
& & \textbf{Furthermore}, \underline{it is unknown whether the role will be played by a male or female actor.} \\
& \multirow{2}{*}{contradiction} & \textbf{There is a rumor that} \textit{The character was announced on 20 September 2019, but casting wasn't revealed until 10 December 2019.} \\
& & \textbf{However, the truth is:} \underline{The character was announced on 19 October 2018.} \\
\bottomrule
\end{tabular}
}
\vspace{-.5em}
\label{tab:negative_gen}
\end{table}

% We present concrete examples of generated texts guided by prompts of different labels in Table~\ref{tab:case_studies}.
% The generated sequences are not only coherent, but also pertain to the corresponding labels.
% For easier tasks like SST-2, the generated texts almost always correctly reflect the desired sentiment polarity specified by the prompt. 
% For more difficult tasks like MNLI, sometimes the generated texts are not of the correct label. 
% (Table~\ref{tab:negative_gen} shows some negative results).
Table~\ref{tab:negative_gen} demonstrates concrete examples where the generated texts may not correctly pertain to the label of the prompt.
The existence of such label noise motivates our use of the regularization techniques in the fine-tuning stage.
% In the future, it will be interesting to develop new methods to better control text generation towards the desired label.
We believe that larger generator PLMs (\eg, GPT-3~\cite{Brown2020LanguageMA}) can bring about better text generation quality and improve the accuracy in producing texts that pertain to the desired class.
Furthermore, better filtering strategies can be developed in the future to select training data with the correct labels.

\end{spacing}

\end{document}